\documentclass{article} 
\usepackage{iclr2024_conference,times}
\pdfoutput=1

\usepackage{amsmath,amsfonts,bm}

\def\1{\bm{1}}

\DeclareMathAlphabet{\mathsfit}{\encodingdefault}{\sfdefault}{m}{sl}
\SetMathAlphabet{\mathsfit}{bold}{\encodingdefault}{\sfdefault}{bx}{n}

\DeclareMathOperator*{\argmax}{arg\,max}

\usepackage{hyperref}
\usepackage{url}
\usepackage{acronym} 
\usepackage{caption}
\usepackage{subcaption}
\usepackage{amsmath}
\usepackage{graphicx}
\usepackage{amssymb}
\usepackage{enumitem}
\usepackage{mathtools}
\usepackage{algorithm}
\usepackage{setspace}
\usepackage[noend]{algpseudocode}
\usepackage{algorithm}
\usepackage{bm}
\usepackage{algorithmicx}
\usepackage{xcolor}
\usepackage{wrapfig}
\usepackage{amsthm}
\usepackage{amssymb}
\renewcommand\cite{\citep}
\definecolor{darkblue}{rgb}{0, 0, 0.5}
\hypersetup{colorlinks=true, citecolor=darkblue, linkcolor=darkblue, urlcolor=darkblue}
\newtheorem{theorem}{Theorem}
\numberwithin{theorem}{section}

\theoremstyle{definition}

\usepackage{multicol}
\usepackage{multirow}
\usepackage{hhline}
\usepackage{array}
\usepackage{booktabs}
\usepackage{bbm}

\acrodef{FWER}{family-wise error rate}
\acrodef{FST}{Fixed Sequence Testing}
\acrodef{SST}{Split Sequence Testing}
\acrodef{SGT}{Sequential Graphical Testing}
\acrodef{MHA}{multi-head attention}
\acrodef{MHT}{multiple hypothesis testing}
\acrodef{LTT}{Learn then Test}
\acrodef{BBO}{Black Box Optimization}
\acrodef{MOO}{multiple objective optimization}
\acrodef{BO}{Bayesian Optimization}
\acrodef{GP}{Gaussian Process}
\acrodef{DDP}{Difference of Demographic Parity}
\acrodef{ERM}{Expected Risk Minimization}
\acrodef{BCE}{Binary Cross-Entropy}
\acrodef{VAE}{Variational Autoencoder}
\acrodef{LHS}{Latin Hypercube Sampling}
\acrodef{HVI}{Hypervolume Improvement}
\acrodef{EHVI}{expected hypervolume improvement}
\acrodef{MOBO}{Multi-Objective-\ac{BO}}

\algtext*{EndWhile}
\algtext*{EndIf}
\algtext*{EndFor}
\algtext*{EndFunction}

\everypar{\looseness=-1}

\usepackage{titlesec}
\titlespacing*{\subsection}{0pt}{.2\baselineskip}{.2\baselineskip}
\titlespacing*{\subsubsection}{0pt}{.2\baselineskip}{.2\baselineskip}
\titlespacing*{\section}{0pt}{.2\baselineskip}{.2\baselineskip}
\title{Risk-Controlling Model Selection via Guided Bayesian Optimization}
\iclrfinalcopy 

\author{Bracha Laufer-Goldshtein$^{1}$\thanks{Correspondence to \texttt{blaufer@tauex.tau.ac.il}} \quad Adam Fisch$^{2}$ \quad \textbf{Regina Barzilay}$^2$ \quad  \textbf{Tommi Jaakkola}$^2$
\\ 
    $^1$Department of EE, Tel-Aviv University \quad $^2$CSAIL, MIT
}

\newcommand{\newpar}[1]{\textbf{{#1}.~}}

\newcommand{\BL}[1]{\textcolor{black}{#1}}
\newcommand{\rev}[1]{\textcolor{black}{#1}}
\newcommand{\red}[1]{\textcolor{red}{#1}}
\newcommand{\blue}[1]{\textcolor{blue}{#1}}

\begin{document}

\maketitle

\begin{abstract}

Adjustable hyperparameters of machine learning models typically impact various key trade-offs such as accuracy, fairness, robustness, or inference cost. Our goal in this paper is to find a configuration that adheres to user-specified limits on certain risks while being useful with respect to other conflicting metrics. We solve this by combining Bayesian Optimization (BO) with rigorous risk-controlling procedures, where our core idea is to steer BO towards an efficient testing strategy. Our BO method identifies a set of Pareto optimal configurations residing in a designated region of interest. The resulting candidates are statistically verified and the best-performing configuration is selected with guaranteed risk levels. We demonstrate the effectiveness of our approach on a range of tasks with multiple desiderata, including low error rates, equitable predictions, handling spurious correlations, managing rate and distortion in generative models, and reducing computational costs. \looseness=-1    
\looseness=-1
\end{abstract}

\section{Introduction}
\label{sec:intro}
Deploying machine learning models in the real-world requires balancing different performance aspects such as low error rate, equality in predictive decisions~\cite{hardt2016equality, pessach2022review}, robustness to spurious correlations~\cite{sagawa2019distributionally, yang2023change}, and model efficiency~\cite{laskaridis2021adaptive,menghani2023efficient}. In many cases, we can influence the model's behavior favorably via sets of hyperparameters that determine the model configuration. However, selecting such a configuration that exactly meets user-defined requirements on test data is typically non-trivial, especially when considering a large number of objectives and configurations that are costly to assess (e.g., that require retraining large neural networks for new settings).\looseness=-1       

\acf{BO} is widely used for efficiently selecting configurations of functions that require expensive evaluation, such as hyperparameters that govern the model architecture or influence the training procedure
~\cite{shahriari2015taking,wang2022recent,bischl2023hyperparameter}. The basic concept is to substitute the costly function of interest with a cheap, and easily optimized, probabilistic surrogate model. This surrogate is used to select promising candidate configurations, while balancing exploration and exploitation. Beyond single-function optimization, \ac{BO} has been extended to multiple objectives, where a set of Pareto optimal configurations that represent the best possible trade-offs is sought~\cite{karl2022multi}. It has also been extended to accommodate multiple inequality constraints~\cite{gardner2014bayesian}. Nevertheless, none of these mechanisms provide formal guarantees on model behavior at test time, and can suffer from unexpected fluctuations from the desired final performance ~\cite{letham2019constrained, feurer2023mind}.\looseness=-1 

Addressing configuration selection from a different prospective, \emph{Learn Then Test} (LTT)~\cite{angelopoulos2021learn} is a rigorous statistical testing framework for controlling multiple risk functions with distribution-free, finite-sample validity in a model-agnostic fashion. Although providing exact theoretical verification, it becomes practically challenging to apply this framework over large configuration spaces due to increased computational costs and loss of statistical power, resulting in the inability to identify useful configurations. These challenges were addressed in the recently proposed \emph{Pareto Testing} method~\cite{laufer2023efficiently}, which combines the complementary features of multi-objective optimization and statistical testing. The core idea is that multi-objective optimization can dramatically reduce the space of configurations to consider, recovering Pareto optimal hyper-parameter combinations that are promising candidates for testing. While improving computational and statistical efficiency, the recovered subspace remains unnecessarily large, containing many irrelevant configurations that are either valid but inefficient or that are highly unlikely to satisfy the constraints. Therefore, when considering expansive configuration spaces, this strategy can again become costly and statistically loose. \looseness=-1          

In this work, we propose a new synergistic approach to combine optimization and testing to achieve efficient model selection under multiple risk constraints. We introduce the notion of the \emph{region of interest} in the objective space that is aligned with the ultimate goal of testing efficiency under limited compute budget.
Our region boundaries are determined by taking into account the data sample sizes and the user-specified limits and certainty levels.   
Consequently, we propose an adjusted \ac{BO} procedure, recovering the part of the Pareto front that intersects with the defined region of interest. The resulting focused optimization procedure recovers a dense set of configurations, representing candidates that are both effective and likely to pass the test. In the final step, we filter the chosen set by means of statistical testing to identify highly-preforming configurations that exhibit verified control.\looseness=-1  

\begin{figure}[!t]
\begin{center}
\includegraphics[width=0.95\textwidth]{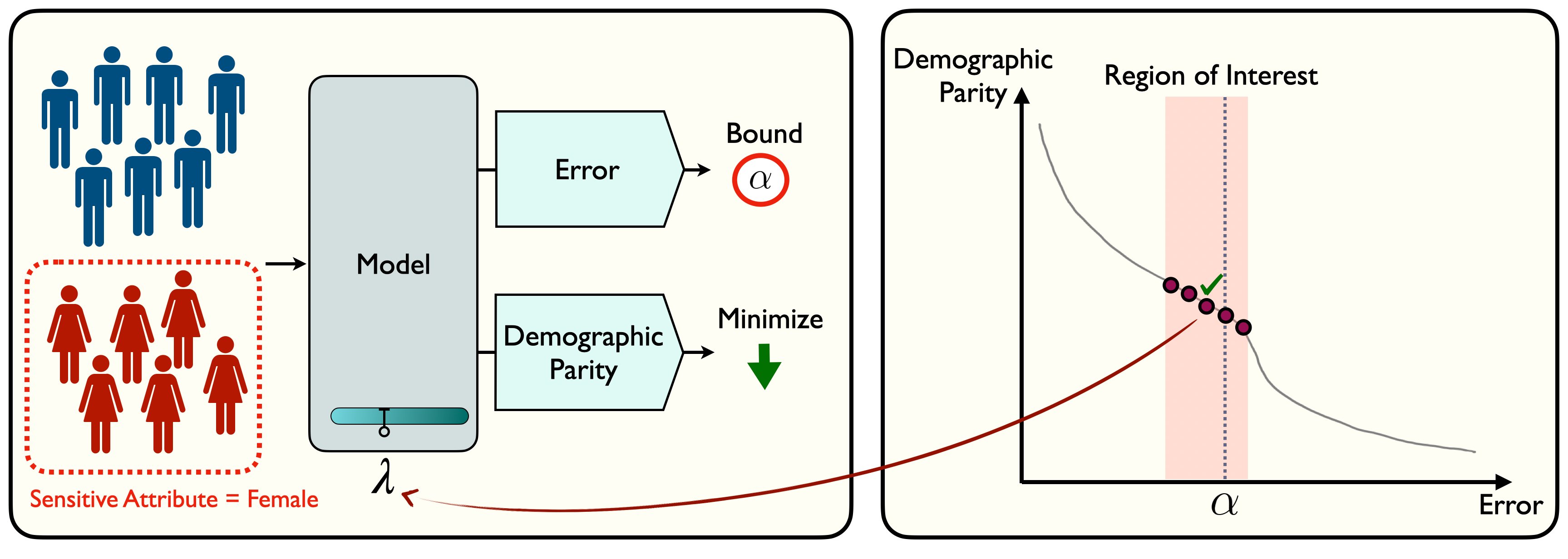}
\end{center}
\vspace{-0.4cm} 
\caption{
\BL{Demonstration of our proposed method for algorithmic fairness with gender as a sensitive attribute (left). We would like to set the model configuration $\lambda$ to minimize the difference in demographic parity, while bounding the overall prediction error by $\alpha$. 
Our method (right): (i) defines a region of interest in the objective space, (ii) identifies Pareto optimal solutions in this region, (iii) statistically validates the chosen solutions, and (iv) sets $\lambda$ to the best-performing verified configuration.}\looseness=-1} 
\label{fig:intro_example}
\vspace{-20pt}
\end{figure}

We show that the proposed framework is flexible and can be applied in diverse contexts for both predictive and generative models, 
and for tuning various types of hyper-parameters that impact the model prior or post training. 
Specifically, we show its applicability in the domains of algorithmic fairness,  robustness to spurious correlations, rate and distortion in \acp{VAE}, and accuracy-cost trade-offs for pruning large-scale Transformer models. See Fig.~\ref{fig:intro_example} for an example and a high-level illustration of the proposed method. \looseness=-1

\textbf{Contribution.} Our main ideas and results can be summarized as follows:
\begin{enumerate}[leftmargin=*,noitemsep]
\item We introduce the region of interest in the objective space that significantly limits the search space for candidate configurations in order to obtain efficient testing outcomes with less computations.
\item We define a new \ac{BO} procedure to identify configurations that are Pareto optimal and lie in the defined region of interest, which are then validated via testing.
\item We present a broad range of objectives across varied tasks, where our approach can be valuable for valid control and effective optimization of diverse performance aspects, including classification fairness, predictive robustness, generation capabilities and model compression.  
\item We demonstrate empirically that our proposed method selects highly efficient and verified configurations under practical budget constraints, relative to baselines. \looseness=-1
\end{enumerate}

\section{Related work}
\label{sec:related}
\newpar{Conformal prediction and risk control} Conformal prediction is a popular model-agnostic and distribution-free uncertainty estimation framework that returns prediction sets or intervals containing the true value with high probability~\cite{vovk2002calibration,vovk2015probabilistic,pmlr-v60-vovk17a, lei-robins-wasserman-dfps, lei2018distribution, gupta2020binary, barber2021predictive}. Coverage validity, provided by standard conformal prediction, has recently been extended to controlling general statistical losses, allowing guarantees in expectation~\cite{angelopoulos2022conformal} or with user-defined probability~\cite{bates2021distribution}. Our contribution builds on the foundational work by~\citet{angelopoulos2021learn} addressing the broader scenario of multiple risk control by selecting a proper low-dimensional hyper-parameter configuration via \ac{MHT}. Additionally, we draw upon the recently introduced Pareto Testing method~\cite{laufer2023efficiently} that further improves computational and statistical efficiency by solving a \ac{MOO} problem and focusing the testing procedure over the approximated Pareto optimal set. In this paper, we point out that recovering the entire Pareto front is redundant and costly and suggest instead to recover a focused part of the front that is aligned with the purpose of efficient testing. This enables highly-expensive hyper-parameter tuning that involves retraining of large models with a limited compute budget. \looseness=-1

\newpar{\acf{BO}} \ac{BO} is a commonly used sequential model-based optimization technique to efficiently find an optimal configuration for a given black-box objective function~\cite{shahriari2015taking,frazier2018tutorial, wang2022recent}. It can be applied to constrained optimization problems~\cite{gardner2014bayesian} or multi-objective scenarios involving several conflicting objectives~\cite{karl2022multi}. However, when used in model hyper-paramaeter tuning, the objective functions can only be approximated through validation data, resulting in no guarantees on test time performance. 
To account for that we resort to statistical testing, and utilize the effectiveness of \ac{BO} to efficiently explore the configuration space and identify promising candidates for testing. Closely related to our work are~\cite{stanton2023bayesian,salinas2023optimizing} proposing to integrate conformal prediction into \ac{BO} in order to improve the optimization process under model misspecification and in the presence of observation noise. These works go in a different direction from our approach, guaranteeing coverage over the approximation of the surrogate model, while ours provides validity on  configuration selection. Another recent work \cite{zhang2023bayesian} utilizes online conformal prediction for maintaining a safety violation rate (limiting the fraction of unsafe configurations found during \ac{BO}), which differs from our provided guarantees and works under the assumption of a Gaussian observation noise. \looseness=-1

\newpar{Multi Objective Optimization (MOO)} Simultaneously optimizing multiple black-box objective functions was traditionally performed with evolutionary algorithms, such as NSGA-II~\cite{deb2002fast}, SMS-EMOA~\cite{emmerich2005emo} and MOEA/D~\cite{zhang2007moea}. Due to the need for numerous evaluations, evolutionary methods can be costly. Alternatively, \ac{BO} methods are more sample efficient and can be combined with evolutionary algorithms. Various methods were proposed exploiting different acquisition functions~\cite{knowles2006parego,belakaria2019max,paria2020flexible} and selection mechanisms, encouraging diversity in the objective space~\cite{belakaria2020uncertainty} or in the design space~\cite{konakovic2020diversity}. The main idea behind our approach is to design a \ac{MOBO} procedure that recovers a small set of configurations that are expected to be both valid and efficient, and then calibrate the chosen set via \ac{MHT}~\cite{angelopoulos2021learn}.\looseness=-1

Additional related work is given in Appendix~\ref{sec:additional_related}.

\section{Problem formulation}
\label{sec:problem}
Consider an input $X\in \mathcal{X}$ and an associated label $Y\in \mathcal{Y}$ drawn from a joint distribution $p_{XY}\in\mathcal{P}_{XY}$. We learn a model $f_{\boldsymbol{\lambda}} \colon \mathcal{X} \rightarrow \mathcal{Y}$, where $\boldsymbol{\lambda}\in\Lambda\subseteq
\mathbb{R}^n$ is an $n$-dimensional hyper-parameter that determines the model configuration. The model weights are optimized over a training set $\mathcal{D}_\textrm{train}$ by minimizing a given loss function, while the hyper-parmeter $\boldsymbol{\lambda}$ determines different aspects of the training procedure or the model final setting. For example, $\boldsymbol{\lambda}$ can weigh the different components of the training loss function, affect the data on which the model is trained, or specify the final mode of operation in a post-processing procedure. \looseness=-1

We wish to select a model configuration $\boldsymbol{\lambda}$ according to different, often conflicting performance aspects, such as low error rate, fairness across different subpopulations and low computational costs. In many practical scenarios, we would like to constrain several of these aspects with pre-specified limits to guarantee a desirable performance in test time. Specifically, we consider a set of objective functions of the form $\ell: \mathcal{P}_{XY}\times\Lambda\rightarrow \mathbb{R}$. \rev{We assume that there are $c$ constrained objective functions $\ell_1, ...., \ell_{c}$, where $\ell_i(\boldsymbol{\lambda})=\mathbb{E}_{p_{XY}}[L_i(f_{\bm{\lambda}}(X),Y,\bm{\lambda})]$ and $L_i:\mathcal{Y}\times \mathcal{Y}\times\Lambda\rightarrow \mathbb{R}$ is a loss function.} In addition, there is a free objective function $\ell_\textrm{free}$ defining a single degree of freedom for minimization. The constraints are specified by the user and have the following form:
\begin{equation}
\label{eq:delta_risk}
\mathbb{P}\left(\ell_i(\boldsymbol{\lambda}) \leq \alpha_i \right) \geq 1 - \delta,\>\> \forall i \in \{1, \ldots, c\}, 
\end{equation}
where $\alpha_i$ is the upper bound of the $i$-th objective function, and $\delta$ is the desired confidence level. The selection is carried out based on two disjoint data subsets: (i) a validation set $\mathcal{D}_\textrm{val}=\{X_i,Y_i\}_{i=1}^k$ and (ii) a calibration set $\mathcal{D}_\textrm{cal}=\{X_i,Y_i\}_{i=k+1}^{k+m}$. We will use the validation data to identify a set of candidate configurations, and the calibration data to validate the identified set. Accordingly, the probability in~\eqref{eq:delta_risk} is defined over the randomness of the calibration data, namely if $\delta=0.1$, then the selected configuration will satisfy the constraints at least $90\%$ of the time across different calibration datasets.\looseness=-1 

We provide here a brief example of our setup in the context of algorithmic fairness and derive other applications in \S\ref{sec:apps}. In many cases, we wish to increase the fairness of the model without significantly sacrificing performance. For example, we would like to encourage similar true positive rates across different subpopulations, while constraining the expected error. One approach to enhance fairness involves introducing a fairness-promoting term in addition to the standard cross-entropy loss~\cite{lohaus2020too,padh2021addressing}. In this case, $\boldsymbol{\lambda}$ represents the weights assigned to each term to determine the overall training loss. Different weights would lead to various accuracy-fairness trade-offs of the resulting model. Our goal is to select a configuration $\boldsymbol{\lambda}$ that optimizes fairness, while guaranteeing with high probability that the overall error would not exceed a certain limit.\looseness=-1

\section{Background}
\label{sec:background} 
In the following, we provide an overview on optimization of multiple objectives and on statistical testing for configuration selection, which are the key components of our method.\looseness=-1

\newpar{Multiple Objective Optimization} Consider an optimization problem over a vector-valued function $\bm{\ell}(\bm{\lambda})=(\ell_1(\bm{\lambda}),\ldots,\ell_{d}(\bm{\lambda}))$ consisting of $d$ objectives. In the case of conflicting objectives, there is no single optimal solution that minimizes them all simultaneously. 
Instead, there is a set of optimal configurations representing different trade-offs of the given objectives. This is the \emph{Pareto optimal set}, defined by:\looseness=-1
\begin{equation}
\Lambda_\textrm{p}=\{\bm{\lambda}\in \Lambda:\;\{\bm{\lambda}^{\prime}\in \Lambda:\;\bm{\lambda}^{\prime }\prec \bm{\lambda},\bm{\lambda}^{\prime }\neq \bm{\lambda}\;\}=\emptyset \},
\label{eq:pareto}
\end{equation}
where $\bm{\lambda}' \prec \bm{\lambda}$ denotes that $\bm{\lambda}'$ \emph{dominates} $\bm{\lambda}$ if for every $i\in \{1,\ldots d\}, \>\ell_i(\bm{\lambda}^{\prime})\leq 
\ell_i(\bm{\lambda})$, and for some $i\in \{1,\ldots d\},\> \ell_i(\bm{\lambda}^{\prime})< 
\ell_i(\bm{\lambda})$. Accordingly, the Pareto optimal set consists of all points that are not dominated by any point in $\Lambda$. 
Given an approximated Pareto front $\hat{\mathcal{P}}$, a  common quality measure is the hypervolume indicator~\cite{zitzler1998multiobjective} 
defined with respect to a \emph{reference point} $\mathbf{r}\in \mathbb{R}^d$:
\begin{equation}
    HV (\hat{\mathcal{P}};\>\mathbf{r}) =\int_{\mathbb{R}^d}\mathbbm{1}_{{H}(\hat{\mathcal{P}},\mathbf{r})}dz
\label{eq:hypervolume}
\end{equation}
where $H(\hat{\mathcal{P}};\mathbf{r})=\{\mathbf{z}\in\mathbb{R}^d:\exists\> \boldsymbol{p}\in\hat{\mathcal{P}}:\mathbf{p}\prec\mathbf{z}\prec\bm{r}\}$ and $\mathbbm{1}_{{H}(\hat{\mathcal{P}},\mathbf{r})}$ is the Dirac delta function that equals 1 if $\mathbf{z}\in H(\hat{\mathcal{P}};\mathbf{r})$ and 0 otherwise. An illustration is provided in Fig.~\ref{fig:hypervolume}. The reference point defines the boundaries for the hypervolume computation. It is usually set to the nadir point that is defined by the worst objective values, so that all Pareto optimal solutions have positive
hypervolume contributions~\cite{ishibuchi2018specify}. For example, in model compression with error and cost as objectives, the reference point can be set to $(1.0, 1.0)$, since the maximum error and the maximum normalized cost equal $1.0$. The hypervolume indicator measures both the individual contribution of each solution to the overall volume,
and the global diversity, reflecting how well the solutions are distributed.
It can be used to evaluate the contribution of a new point to the current approximation, defined as the \ac{HVI}:
\begin{equation}
HVI(\bm{\ell}(\bm{\lambda}),\hat{\mathcal{P}};\mathbf{r})=HV(\bm{\ell}(\bm{\lambda})\cup\hat{\mathcal{P}};\>\mathbf{r})  - HV(\hat{\mathcal{P}};\>\mathbf{r}).
\label{eq:hvi}
\end{equation}
The hypervolume indicator serves both as a performance measure for comparing different algorithms and as a score for maximization in various MOO methods~\cite{emmerich2005emo, emmerich2006single,bader2011hype,daulton2021parallel}.\looseness=-1 

\newpar{\ac{BO}} \ac{BO} is a powerful tool for optimizing black-box objective functions that are expensive to evaluate. It uses a \emph{surrogate model} to approximate the expensive objective function, and iteratively selects new points for evaluation based on an \emph{acquisition function} that balances exploration and exploitation. Formally, we start with an initial pool of random configurations $\mathcal{C}_{0}=\{\bm{\lambda}_0,\ldots,\bm{\lambda}_{N_0}\}$ and their associated objective values $\mathcal{L}_{0}=\{\ell(\bm{\lambda}_1),\ldots,\ell(\bm{\lambda}_{N_0})\}$. 
Commonly, a \ac{GP}~\cite{williams2006gaussian} serves as a surrogate model, providing an estimate with uncertainty given by the Gaussian posterior. 
We assume a zero-mean \ac{GP} prior $g(\bm{\lambda})\sim\mathcal{N}\left(0,k(\bm{\lambda},\bm{\lambda})\right)$, characterized by a kernel function $\kappa:\Lambda\times \Lambda\rightarrow \mathbb{R}$. The posterior distribution of the \ac{GP} is given by $p(g|\bm{\lambda},\mathcal{C}_{n},\mathcal{L}_n)=\mathcal{N}\left(\mu(\bm{\lambda}),\Sigma(\bm{\lambda},\bm{\lambda})\right)$, with $\mu(\bm{\lambda})=\mathbf{k}(\mathbf{K}+\sigma^2\mathbf{I})^{-1}\mathbf{l}$ and $\Sigma(\bm{\lambda},\bm{\lambda})=k(\bm{\lambda},\bm{\lambda})-\mathbf{k}^T\left(\mathbf{K}+\sigma^2\mathbf{I}\right)^{-1}\mathbf{k}$, where $k_i=\kappa(\bm{\lambda},\bm{\lambda}_i), \> K_{ij}=\kappa(\bm{\lambda}_i,\bm{\lambda}_j)$ and $l_i=\ell(\bm{\lambda}_i), i,j\in\{1,\ldots,|\mathcal{C}_n|\}$. Here $\sigma^2$ is the observation noise variance, i.e. $\ell(\bm{\lambda}_i)\sim\mathcal{N}(g(\bm{\lambda}_i),\sigma^2)$.
Next, we optimize an acquisition function that is defined on top of the surrogate model, such as probability of improvement (PI)~\cite{kushner1964new}, expected improvement (EI)~\cite{movckus1975bayesian}, and lower confidence bound (LCB)~\cite{auer2002using}. For multi-objective optimization, a GP is fitted to each objective. Then, one approach is to perform scalarization~\cite{knowles2006parego}, converting the problem back to single-objective optimization and applying one of the aforementioned acquisition functions. Another option is to use a modified acquisition function that is specified for the multi-objective case, such as \ac{EHVI}~\cite{emmerich2006single} and predictive entropy search for multi-objective optimization (PESMO)~\cite{hernandez2016predictive}. After a new configuration is selected, it is evaluated and added to the pull. This process is repeated until the maximum number of iterations is reached.\looseness=-1     

\newpar{\acf{LTT} \& Pareto Testing} \citet{angelopoulos2021learn} have recently proposed \ac{LTT}, which is a statistical  framework for configuration selection based on \ac{MHT}. Given a set of constraints of the form~\eqref{eq:delta_risk}, a null hypothesis is defined as $H_{\bm{\lambda}}: \exists ~i \text{ where } \ell_i(\bm{\lambda}) > \alpha_i$ i.e., that at least one of the constraints is \emph{not} satisfied. For a given configuration, we can compute the p-value under the null-hypothesis based on the calibration data. If the p-value is lower than the significance level $\delta$, the null hypothesis is rejected and the configuration is declared to be valid. When testing multiple model configurations simultaneously, this becomes an \ac{MHT} problem. In this case, it is necessary to apply a correction procedure to control the \ac{FWER}, i.e. to ensure that the probability of one or more wrong rejections is bounded by $\delta$. 
This can become computationally demanding and result in inefficient testing when the configuration space is large. In order to mitigate these challenges, Pareto Testing was proposed~\cite{laufer2023efficiently}, where the testing is focused on the most promising configurations identified using \ac{MOO}. Accordingly, only Pareto optimal configurations are considered and are ranked by their approximated p-values from low to high risk. Then, \ac{FST}~\cite{holm1979simple} is applied over the ordered set, sequentially testing the configurations with a fixed threshold $\delta$ until failing to reject for the first time. 
Although Pareto Testing demonstrates enhanced testing efficiency, it recovers the entire Pareto front, albeit focusing only on a small portion of it during testing.
Consequently, the optimization budget is not directly utilized in a way that enhances testing efficiency, putting an emphasis on irrelevant configurations on one side and facing an excessive sparsity within the relevant area on the other.\looseness=-1

\section{Method}
\label{sec:pareto_testing}
Our approach involves two main steps: (i) performing \ac{BO} to generate a small set of potential configurations, and (ii) applying \ac{MHT} over the candidate set to identify valid configurations.  
Considering the shortcomings of Pareto Testing, we argue that the two disjoint stages of optimization followed by testing are suboptimal, especially for resource-intensive \ac{MOO}.
As an alternative, we propose adjusting the optimization procedure for better testing outcomes by focusing only on the most relevant parts in the objective space. To accomplish this, we need to (i) specify a \emph{region of interest} guided by our testing goal, and (ii) establish a \ac{BO} procedure capable of effectively identifying configurations within the defined region. In the following we describe these steps in details.\looseness=-1 

\subsection{Defining the Region of Interest}
We would like to define a region of interest in the objective space $\mathbb{R}^{c+1}$, where we wish to identify candidate configurations that are likely to be valid and efficient while conducting \ac{MHT}. We start with the case of a single constraint ($c=1$). Recall that in the testing stage we define the null hypothesis $H_{\bm{\lambda}}:  \ell(\bm{\lambda}) > \alpha$ for a candidate configuration $\bm{\lambda}$, and compute a p-value for a given empirical loss over the calibration data $\hat{\ell}(\bm{\lambda})=\frac{1}{m}\sum_{j=k+1}^{k+m} \ell(X_j,Y_j;\bm{\lambda})$. \rev{A valid p-value $p_{\bm{\lambda}}$ has to be super-uniform under the null hypothesis, i.e. $\mathbb{P}\left(p_{\bm{\lambda}} \leq u\right) \leq u$, for all $u\in[0,1]$. As presented in~\cite{angelopoulos2021learn}, a valid p-value can be computed based on concentration inequalities that quantify how close is the sample loss to the expected population loss. When the loss is bounded by $1$, we can use Hoeffding's inequality to obtain the following p-value (see Appendix~\ref{sec:math_details}):}\looseness=-1 
\begin{equation}
    p_{\bm{\lambda}}^\textrm{HF}\coloneqq e^{-2 m\left(\alpha-\hat{\ell}(\bm{\lambda})\right)^2_+}.
    \label{eq:p_value}
\end{equation}
\rev{For a given significance level $\delta$, the null hypothesis is rejected (the configuration is declared to be risk-condoling), when $p_{\bm{\lambda}}^\textrm{HF}<\delta$. By rearranging~\eqref{eq:p_value}, we obtain that the maximum empirical loss $\hat{\ell}(\bm{\lambda})$ that can pass the test with significance level $\delta$ is given by (see Appendix~\ref{sec:math_details}):}\looseness=-1 
\begin{equation}
\alpha^\textrm{max}=\alpha-\sqrt{\frac{\log\left(1/\delta\right)}{2m}}.
\label{eq:alpha_max}
\end{equation}
\rev{For example, consider the error rate as a loss function, which we would like to bound by $5\%$ ($\alpha=0.05$), with significance level $\delta=0.1$. By~\eqref{eq:alpha_max}, if the empirical loss of a calibration set of size $m=5000$ is up to $3.5\%$, then we have enough evidence to declare that this configuration is safe and its error does not exceed $5\%$.}

\rev{In the \ac{BO} procedure, we are interested in identifying configurations that are likely to be both valid and efficient. 
On the one hand, in order to be valid the loss must not exceed $\alpha^\textrm{max}$. On the other hand, from efficiency considerations, we would like to minimize the free objective as much as possible. This means that the constrained loss should be close to $\alpha^\textrm{max}$ (from bellow), since the free objective decreases as the constrained objective increases. An illustration demonstrating this idea is provided in~Fig.~\ref{fig:explain}, where the irrelevant regions are: (i) the green part on the left where the configurations are not effectively minimizing $\ell_2$, and (2) the brown part on the right where the configurations are not satisfying the constraint. Ideally, we would like to find configurations with expected loss equal to the limiting testing threshold $\alpha^\textrm{max}$. However, during optimization we can only evaluate the loss on a finite-size validation data with $|\mathcal{D}_\textrm{val}|=k$. To account for that, we construct an interval $[\ell^\textrm{low},\ell^\textrm{high}]$ around $\alpha^\textrm{max}$ based on the size of the validation data. In this region, we wish to include empirical loss values that are \emph{likely} to correspond to an
expected value of $\alpha^\textrm{max}$ based on the evidence provided by the validation data. 
Specifically, we consider $\hat{\ell}_1$ values that are likely to be obtained under $\ell_1=\alpha^\textrm{max}$ with probability that is at least $\delta'$. 
This can be formed by defining $1-\delta'$ confidence bounds. For example, using again Hoeffding's inequality, we obtain the following region of interest:} \looseness=-1 
\begin{equation}
R(\alpha,k,m)=\left[ \underbrace{\alpha^\textrm{max}-\sqrt{\frac{\log\left(1/\delta'\right)}{2k}}}_{\ell^\textrm{low}}, \underbrace{\alpha^\textrm{max}+\sqrt{\frac{\log\left(1/\delta'\right)}{2k}}}_{\ell^\textrm{high}}\right]. 
\label{eq:region_i}
\end{equation}
\rev{Note that setting the value of $\delta'$ is an empirical choice that is unrelated to the \ac{MHT} procedure and to $\delta$. For small $\delta'$ the region expands, including more options with reduced density, while for larger $\delta'$ the region becomes smaller and denser. In any case, when $k$ increases, the width of~\eqref{eq:region_i} decreases as we have more confidence in the observed empirical losses of being reflective of the expected loss.} In practice we use the tighter Hoeffding-Bentkus inequality for both~\eqref{eq:alpha_max} and~\eqref{eq:region_i} (see Appendix~\ref{sec:math_details}).

In the case of multiple constraints, the null hypothesis is defined as $H_{\bm{\lambda}}: \exists ~i \text{ where } \ell_i(\bm{\lambda}) > \alpha_i$. A valid p-value is given by $p_{\bm{\lambda}} = \max_{i\in\{1,\ldots,c\}} p_{\bm{\lambda},i}$, where $p_{\bm{\lambda},i}$ is the p-value corresponding to the $i$-th constraint. Consequently, we define the region of interest in the multi-constraint case as the intersection of the individual regions:\looseness=-1 
\begin{equation}
R(\bm{\alpha},k,m)=\bigcap_{i=1}^c R(\alpha_i,k,m);\>\>\> \bm{\alpha}=(\alpha_1,\ldots,\alpha_c)
\label{eq:region}
\end{equation}

\subsection{Local Hypervolume Improvement}

\begin{wrapfigure}{h}{0.35\textwidth}
\vspace*{-11ex}
\begin{center}
\includegraphics[width=0.35\textwidth]{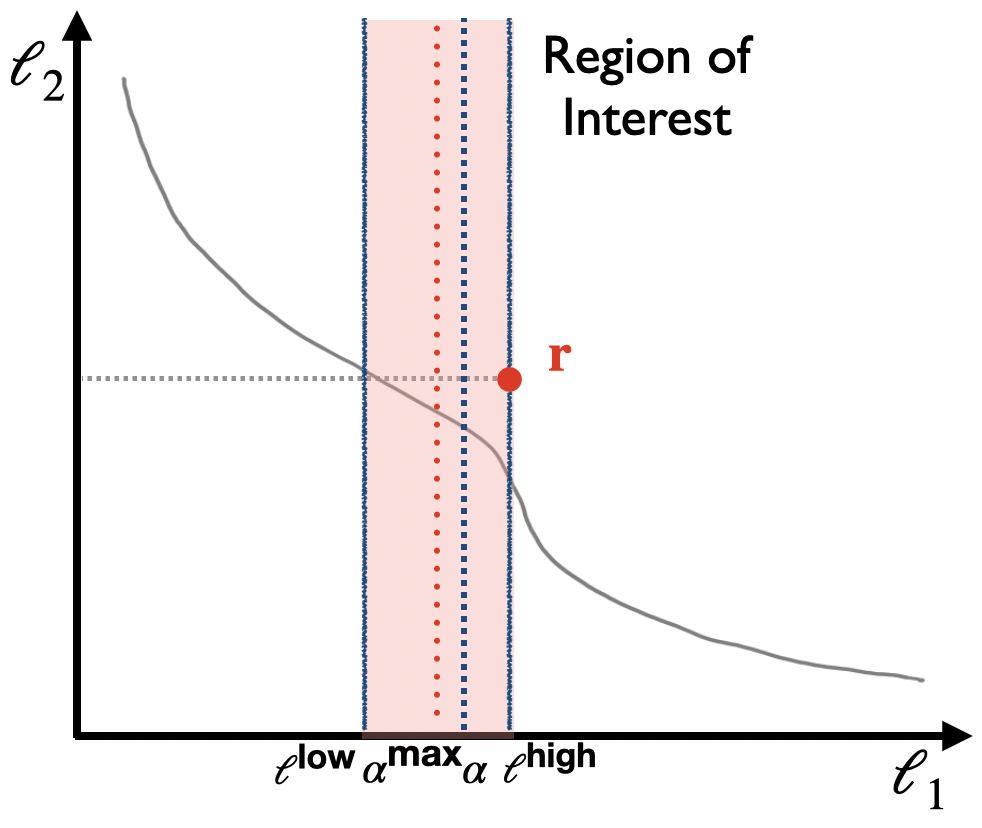}
  \end{center}
\vspace*{-4ex}
  \caption{Proposed \ac{BO} procedure for two objectives. $\ell_1$ is controlled at $\alpha$ while $\ell_2$ is minimized. The shaded area corresponds to our defined region of interest. A reference point (in red) is defined accordingly to enclose 
  the region of interest. \looseness=-1}
\vspace*{-6.5ex}
  \label{fig:bo_illustartion}
\end{wrapfigure}

Given our definition of the region of interest, we derive a \ac{BO} procedure that recovers Pareto optimal points in the intersection of $R(\bm{\alpha},k,m)$ and $\mathcal{P}$. Our key idea is to use the \ac{HVI} in~\eqref{eq:hvi} as an acquisition function and to modify it to capture only the region of interest. To this end, we properly define the reference point $\mathbf{r}\in \mathbb{R}^{c+1}$ to enclose the desired region. \looseness=-1

Recall that the reference point defines the upper limit in each direction. Therefore, we set $r_i=\ell_i^\textrm{high}, \> i\in\{1,\ldots,c\}$ using the upper bound in~\eqref{eq:region_i} for the constrained dimensions. We can use the maximum possible value of $\ell_\textrm{free}$ for $r_{c+1}$. However, this will unnecessarily increase the defined region, including configurations that are low-risk but do not minimize the free objective (where the constrained objectives are overly small and the free objective is overly big).
Instead, we set $r_{c+1}$ to be the point on the free axis that correspond to the intersection of the lower limits of the constrained dimensions. For this purpose, we use the posterior mean as our objective estimator, i.e. $\hat{g} =\mu$. We define the region $R^\textrm{low}=\left\{\boldsymbol{\lambda}:\hat{g}_1(\boldsymbol{\lambda})<\ell_1^\textrm{low},\ldots,\hat{g}_c(\boldsymbol{\lambda})<\ell_c^\textrm{low}\right\}$, where the configurations are likely to be valid but inefficient. Finally, we tightly enclose this region from below in the free dimension:\looseness=-1
\begin{equation}
r_{c+1} = \min_{\boldsymbol{\lambda}\in R^\textrm{low}}\hat{g}_\textrm{free}(\boldsymbol{\lambda}).   
\end{equation}
As a result, we obtain the following reference point:
\begin{equation}
\mathbf{r} = \left(\ell_1^\textrm{high},\ldots,\ell_c^\textrm{high},\> \min_{\boldsymbol{\lambda}\in R^\textrm{low}}\hat{g}_\textrm{free}(\boldsymbol{\lambda})\right).
\label{eq:ref_point}
\end{equation}
We select the next configuration by maximizing the \ac{HVI}~\eqref{eq:hvi} with respect to this reference point: \looseness=-1 
\begin{equation}
\bm{\lambda}_n=\argmax_{\bm{\lambda}}HVI(\hat{\bm{g}}(\bm{\lambda}),\hat{\mathcal{P}};\mathbf{r}) 
\end{equation}
to recover only the relevant section and not the entire Pareto front.
We evaluate the objective functions on the new selected configuration, and update our candidate set accordingly. This process of \ac{BO} iterations continues until reaching the maximum budget $N$. The resulting candidate set is denoted as $\mathcal{C}^{BO}$. Our proposed \ac{BO} procedure is described in Algorithm~\ref{alg:bo} and is illustrated in Fig.~\ref{fig:bo_illustartion}.\looseness=-1  

Note that in \ac{MOBO} it is common to use an \ac{HVI}-based acquisition function that also takes into account the predictive uncertainty as in EHVI~\cite{emmerich2005emo} and SMS-EGO~\cite{ponweiser2008multiobjective}.
However, our preliminary runs showed that these approaches do not work well in the examined scenarios with small budget ($N\in[10,50]$), as they often generated points outside the region of interest. Similarly, for these scenarios the random scalarization approach, proposed in~\cite{paria2020flexible}, was less effective for generating well-distributed points inside the desired region.\looseness=-1 

\subsection{Testing the Final Selection}
We follow~\cite{angelopoulos2021learn,laufer2023efficiently} for testing the selected set. Prior to testing we filter and order the candidate set $\mathcal{C}^\textrm{BO}$. Specifically, we retain only Pareto optimal configurations from $\mathcal{C}^\textrm{BO}$, and arrange the remaining configurations by increasing p-values (approximated by $\mathcal{D}_\textrm{val}$). Next, we recompute the p-values based on $\mathcal{D}_\textrm{cal}$ and perform \ac{FST}, where we start testing from the first configuration and continue until the first time the p-value exceeds $\delta$. As a result, we obtain the validated set $\mathcal{C}^\textrm{valid}$, and choose a configuration minimizing the free objective:
\begin{equation}
\bm{\lambda}^*=\min_{\bm{\lambda}\in \mathcal{C}^\textrm{valid}}\ell_{\textrm{free}}(\bm{\lambda}).
\label{eq:select}
\end{equation} 
Our method is summarized in Algorithm~\ref{alg:selection}. As a consequence of~\cite{angelopoulos2021learn, laufer2023efficiently} we achieve a valid risk-controlling configuration, as we now formally state.\looseness=-1 

\begin{theorem}
\label{thm:ltt}
Let $\mathcal{D}_\mathrm{val}=\{X_i,Y_i\}_{i=1}^k$ and $\mathcal{D}_\mathrm{cal}=\{X_i,Y_i\}_{i=k+1}^{k+m}$ be two disjoint datasets. 
Suppose the p-value $p_{\bm{\lambda}}$, derived from $\mathcal{D}_\mathrm{cal}$, is super-uniform under $\mathcal{H}_{\bm{\lambda}}$ for all ${\bm{\lambda}}$. Then the output $\bm{\lambda}^*$ of Algorithm~\ref{alg:selection} satisfies Eq.~\eqref{eq:delta_risk}.
\end{theorem}

In situations where we are unable to identify any statistically valid configuration (i.e., $\mathcal{C}^\textrm{valid} = \emptyset$), we set $\bm{\lambda} = \texttt{null}$. In practice, the user can choose limits $\alpha_1,\ldots,\alpha_c$ that are likely to be feasible based on the initial pool of configurations $\mathcal{C}_0$ that is generated at the beginning of the \ac{BO} procedure. Specifically, the user may select $\alpha_i\in [\min_{\bm{\lambda}\in \mathcal{C}_0}\ell_i(\bm{\lambda}), \max_{\bm{\lambda}\in \mathcal{C}_0}\ell_i(\bm{\lambda})], i\in\{1,\ldots,c\}$, and can further refine this choice during the \ac{BO} iterations as more function evaluations are accumulated.  

\section{Applications}
\label{sec:apps}
We demonstrate the effectiveness of our proposed method for different tasks with diverse objectives, where the definition of $\bm{\lambda}$ and its effect prior or post training, vary per setting.   

\newpar{Classification Fairness} In many classification tasks, it is important to take into account the behavior of the predictor with respect to different subpopulations.
Assuming a binary classification task and a binary sensitive attribute $a=\{-1,1\}$, we consider the \ac{DDP} as a fairness score~\cite{wu2019convexity}:
\begin{equation}
\textrm{DDP}(f) = \mathbb{E}\left[\mathbbm{1}_{f(x)>0}|a=-1\right] 
-\mathbb{E}\left[\mathbbm{1}_{f(x)>0}|a=1\right].
\label{eq:ddp}
\end{equation}
We define the following loss parameterized by $\lambda$:
\begin{equation}
R(f;\lambda) = (1-\lambda) \cdot \textrm{BCE}(f) +\lambda \cdot\widehat{\textrm{DDP}}(f), 
\end{equation}
 where $\textrm{BCE}$ is the binary cross-entropy loss, and $\widehat{\textrm{DDP}}$ is the hyperbolic tangent relaxation of~\eqref{eq:ddp}~\cite{padh2021addressing}. Changing the value of $\lambda$ leads to different models that trade-off accuracy for fairness. In this setup, we have a 1-dimensional hyperparamter $\lambda$ and two objectives: (i) the error of the model $\ell_\textrm{err}(\lambda)=\mathbb{E}\left[\mathbbm{1}_{f_\lambda(X) \neq Y}\right]$, and (ii) the \ac{DDP} defined in~\eqref{eq:ddp} $\ell_\textrm{ddp}(\lambda)=\textrm{DDP}(f_\lambda)$.\looseness=-1 

\newpar{Classification Robustness} Predictors often rely on spurious correlations found in the data (such as background features), which leads to significant performance variations among different subgroups. Recently, \citet{izmailov2022feature} demonstrated that models trained using expected risk minimization surprisingly learn core features in addition to spurious ones.
Accordingly, they proposed to enhance model robustness by retraining the final layer on a balanced dataset. We adapt their approach to obtain different configurations, offering a trade-off between robustness and average performance.\looseness=-1  

Given a dataset $\mathcal{D}$ (either the training set or a part of the validation set) we denote by $\mathcal{D}_b$ a balanced subset of $\mathcal{D}$ with equal number of samples per subgroup, and by $\mathcal{D}_u$ a random (unbalanced) subset of $\mathcal{D}$. We define a parameterized dataset $\mathcal{D}_{\lambda}$ in the following way.
Let $B\sim\textrm{Bern}(\lambda)$ denote a Bernoulli random variable with parameter $\lambda$. We randomly draw $K$ i.i.d samples $\{B_i\}_{i=1}^K$, and construct $\mathcal{D}_\lambda=\{X_i,Y_i\}_{i=1}^K$, where $(X_i,Y_i)$ are randomly drawn from $\mathcal{D}_b$ if $B_i=1$ or from $\mathcal{D}_u$, otherwise. We train the last layer with binary cross-entropy loss on the resulting dataset $\mathcal{D}_\lambda$. 
As a result, we have a 1-dimensional hyper-parameter $\lambda$ that controls the degree to which the dataset is balanced. We define two objective functions: (i) the average error $\ell_\textrm{err}(\lambda)=\mathbb{E}\left[\mathbbm{1}_{f_\lambda(X) \neq Y}\right]$, and (ii) the worst error over all subgroups $\ell_\textrm{worst-err}(\lambda)=\max_{g\in\mathcal{G}}\mathbb{E}\left[\mathbbm{1}_{f_\lambda(X) \neq Y}|G=g\right]$ where $G\in\mathcal{G}$ is the group label.\looseness=-1 

We also examine the case of \emph{selective} classification and robustness. The selective classifier can abstain from making a prediction when the confidence is lower then a threshold $\tau$, i.e. $f_\lambda(x)<\tau$. In this case, we have a 2-dimensional hyper-parmeter  $\bm{\lambda}=(\lambda,\tau)$ and an additional objective function of the mis-coverage rate (where the predictor decides to abstain) 
$\ell_\textrm{mis-cover}(\lambda)=\mathbb{E}\left[\mathbbm{1}_{f_\lambda(x)<\tau}\right]$.\looseness=-1 

\newpar{\ac{VAE}} \acfp{VAE}~\cite{kingma2013auto,rezende2014stochastic} are generative models that leverage a variational approach to learn the latent variables underlying the data, and can generate new samples by sampling from the learned latent space.  We focus on a $\beta$-\ac{VAE}~\cite{higgins2016beta}, which balances the reconstruction error (distortion) and the KL divergence (rate):
\begin{equation}
R(f;\beta)=\mathbb{E}_{p_d(x)}\left[\mathbb{E}_{q_\phi(z|x)}\left[-\log p_\theta(x|z)\right]\right]+\beta\cdot\mathbb{E}_{p_d(x)}\left[D_{KL}(q_\phi(z|x)||p(z))\right],
\label{eq:vae}
\end{equation}
where $f$ consists of an encoder $q_\phi(z|x)$ and a decoder $p_\theta(x|z)$, parameterized by $\phi$ and $\theta$, respectively,  and $p(z)$ is the latent prior distribution. Generally, models with low distortion perform high-quality reconstruction but generate less realistic samples and vice versa. We have a single parameter $\lambda=\beta$ and two objectives $\ell_\textrm{recon}(f)$, $\ell_\textrm{KLD}(f)$ defined by the left and right terms in~\eqref{eq:vae}, respectively.\looseness=-1 

\newpar{Transformer Pruning} We adopt the three dimensional transformer pruning scheme proposed in~\cite{laufer2023efficiently}: (i) token pruning, removing unimportant tokens from the input sequence, (ii) layer early-exiting, computing part of the model's layers for easy examples, and (iii) head pruning, removing attention heads from the model architecture. We obtain $\bm{\lambda}=(\lambda_1,\lambda_2,\lambda_3)$ with the three thresholds controlling the pruning in each dimension, and consider two objectives: (i) the accuracy difference between the pruned model and the full model $\ell_\textrm{diff-acc}(\lambda)=\mathbb{E}\left[\mathbbm{1}_{f(X) = Y}-\mathbbm{1}_{f_{\bm{\lambda}}(X) = Y}\right]$ and (ii) the respective cost ratio $\ell_\textrm{cost}(\lambda)=\mathbb{E}\left[\frac{C(f_{\bm{\lambda}}(X))}{C(f(X))}\right]$.\looseness=-1 
 
\section{Experiments}
\label{sec:results}
We briefly describe the experimental setup and present our main results. Detailed setup information, as well as additional results are provided in Appendixes~\ref{sec:exp_details} and~\ref{sec:more_res}, respectively.\looseness=-1 

\newpar{Baselines} 
\rev{We compare the proposed method to other baselines that differ only in the first stage by their optimization mechanism. The second testing stage is the same for all baselines (and the proposed method), therefore all baselines can be considered as variants of Pareto Testing~\cite{laufer2023efficiently}. We define two simple baselines:
\textsc{uniform} - a uniform grid in the hyper-parameter space.
\textsc{Random} - a uniform random sampling for $n=1$ and Latin Hypercube Sampling (LHS)~\cite{mckay2000comparison} for $n>1$.
In addition, we compare to multi-objective optimizers:
\textsc{HVI} (same acquisition function as in the proposed method) and \textsc{EHVI} ~\cite{emmerich2006single} with reference point defined by the maximum loss values, and 
\textsc{ParEGO} ~\citep{knowles2006parego, cristescu2015surrogate} using \textsc{Smac3} implementation~\citep{lindauer2022smac3}. 
We choose the values of $\alpha$ for each task according to the range obtained from the initial pool of configurations. See table~\ref{tab:dataset} for the range values for each objective. We set $\delta=0.1$ and $\delta'=0.0001$.} \looseness=-1  

\newpar{Datasets} We use the following datasets: \textbf{Fairness - Adult}~\cite{dua2017uci}, predict if the income is above $50$k\$ with gender as a sensitive attribute; 
\textbf{Robustness 
 - CelebA}~\cite{lin2019pareto}, predict if a person has blond hair, where the spurious correlation is the gender; \textbf{VAE - MNIST}~\cite{lecun1998mnist}; \textbf{Pruning - AG News}~\cite{zhang2015character}, 
topic news classification.\looseness=-1 

\begin{figure}[t]
\begin{center}
\begin{subfigure}[b]{0.25\textwidth}
 \centering
 \includegraphics[width=\textwidth]{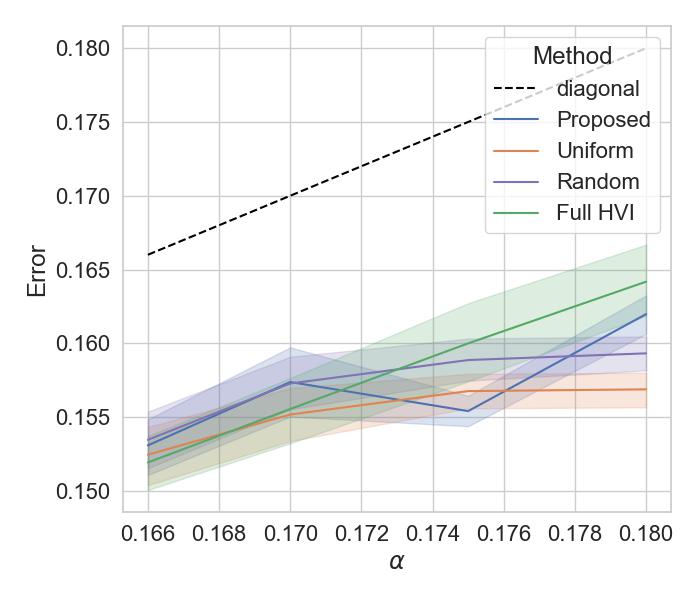}
\end{subfigure}
\vspace{-0.15em}
\hspace{-0.5em}
\begin{subfigure}[b]{0.25\textwidth}
 \centering
 \includegraphics[width=\textwidth]{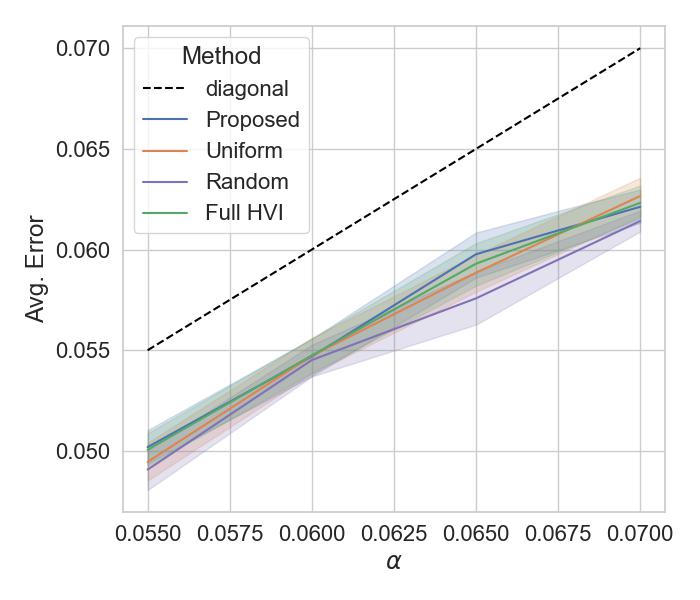}
\end{subfigure}
\vspace{-0.15em}
\hspace{-0.5em}
\begin{subfigure}[b]{0.25\textwidth}
 \centering
 \includegraphics[width=\textwidth]{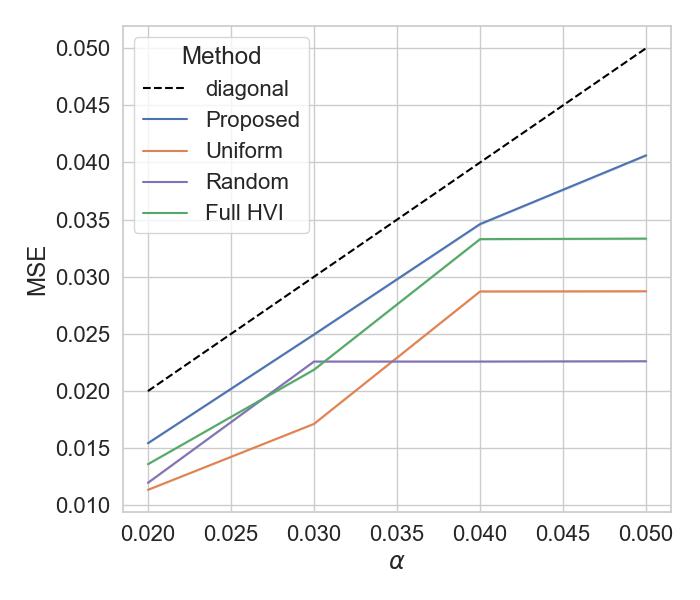}
\end{subfigure}
\vspace{-0.15em}
\hspace{-0.5em}
\begin{subfigure}[b]{0.25\textwidth}
 \centering
 \includegraphics[width=\textwidth]{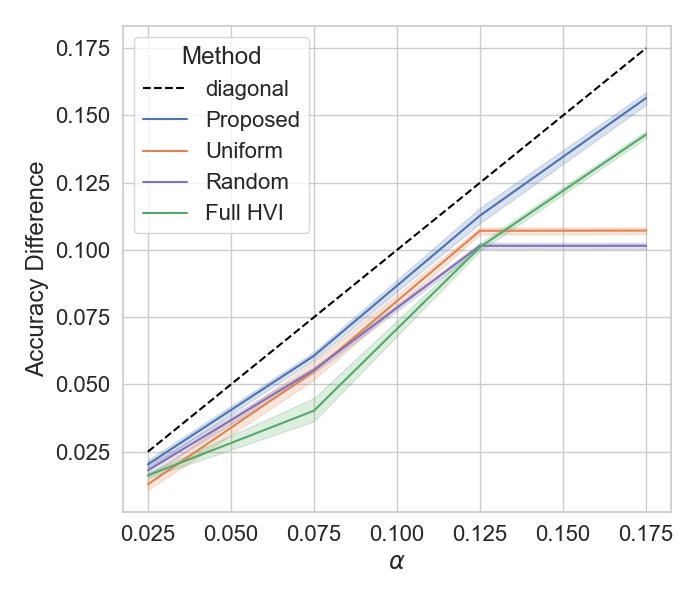}
\end{subfigure}
\vspace{-0.15em}
\begin{subfigure}[b]{0.25\textwidth}
 \centering
 \includegraphics[width=\textwidth]{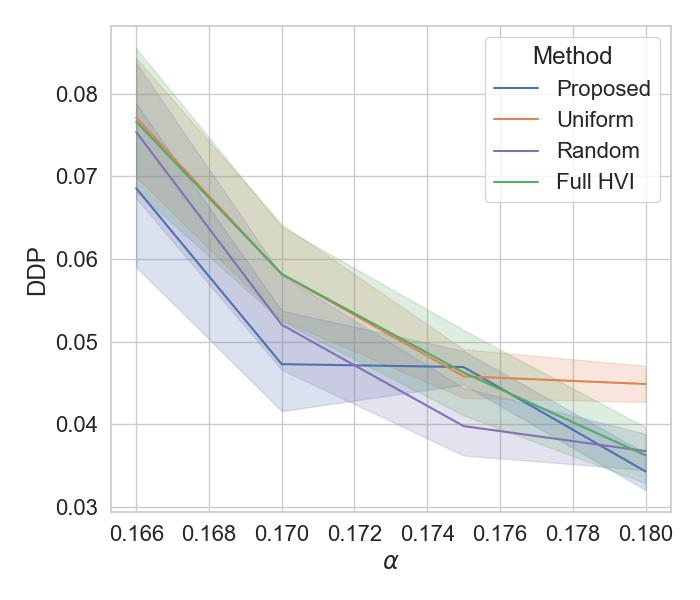}
\vspace{-1em}
\caption{Fairness}
\end{subfigure}
\hspace{-0.5em}
\begin{subfigure}[b]{0.25\textwidth}
 \centering
 \includegraphics[width=\textwidth]{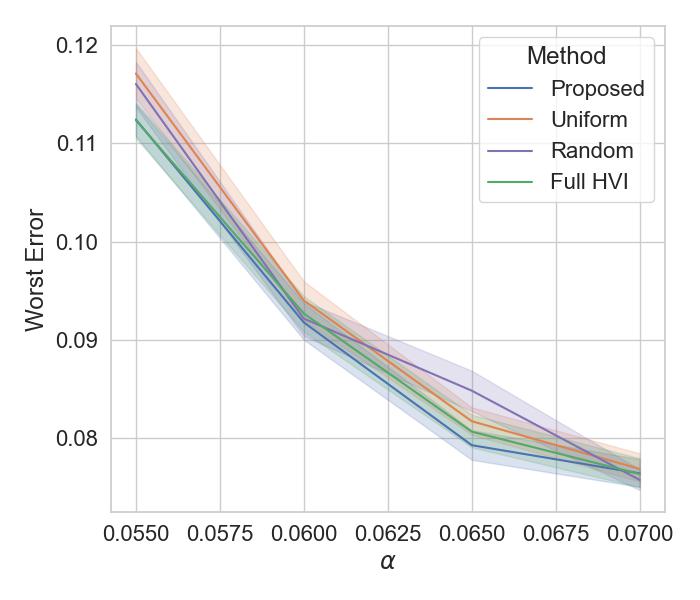}
\vspace{-1em}
\caption{Robustness}
\end{subfigure}
\hspace{-0.5em}
\begin{subfigure}[b]{0.25\textwidth}
 \centering
 \includegraphics[width=\textwidth]{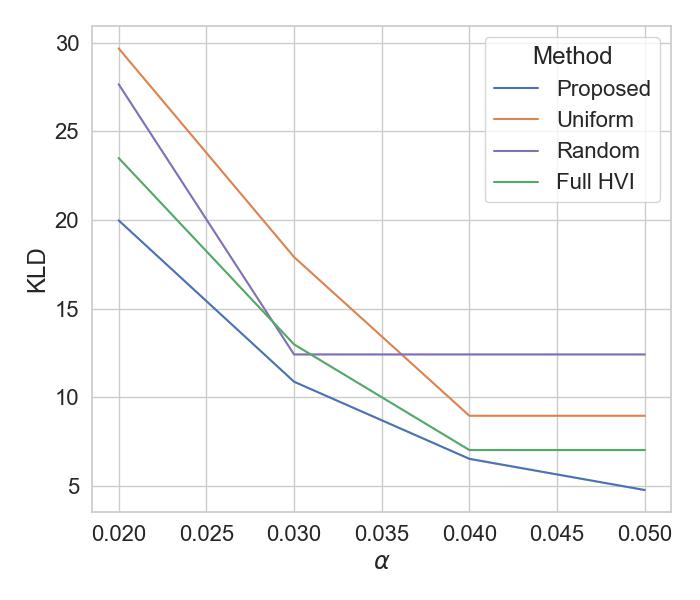}
\vspace{-1em}
 \caption{VAE}
 \end{subfigure}
\hspace{-0.5em}
\begin{subfigure}[b]{0.25\textwidth}
 \centering
 \includegraphics[width=\textwidth]{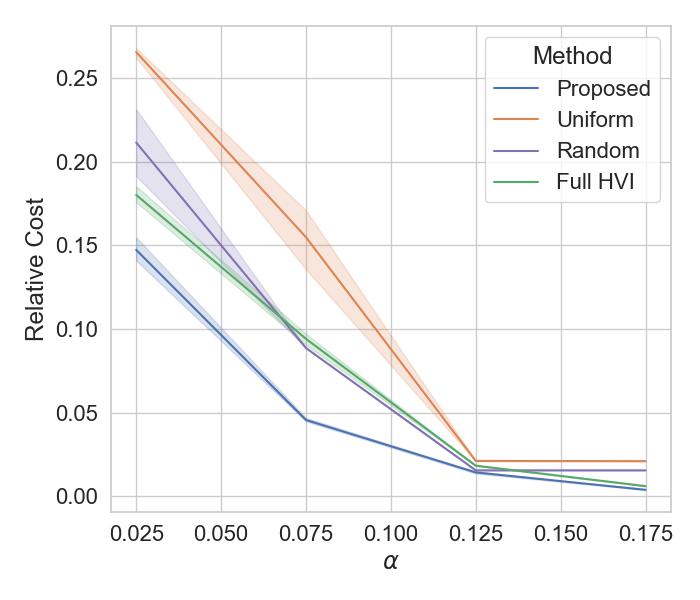}
\vspace{-0.8em} 
  \caption{Pruning}
\end{subfigure}
\end{center}
\vspace{-10pt} 
\caption{\rev{Two objectives. 
Presents constrained (top) and free objectives (bottom). \textsc{ParEGO} and \textsc{EHVI} baselines appear in Fig.~\ref{fig:res_contorol_two_add} for the sake of clarity.}\looseness=-1} 
\label{fig:res_control_two}
\vspace{-25pt} 
\end{figure}

\newpar{Two objectives}
We examine the following scenarios: \textbf{Fairness} - error is controlled and \ac{DDP} is minimized; \textbf{Robustness} - avg. error is controlled and worst error is minimized; \textbf{VAE} - reconstruction error is controlled and KLD is minimized; \textbf{Pruning} - error difference is controlled and relative cost is minimized. \rev{Results are presented in Figs.~\ref{fig:res_control_two} and~\ref{fig:res_contorol_two_add} , showing the mean scores over $50$ random calibration and test splits. Shaded regions correspond to 95\% CI. We see that the proposed method is superior over all baselines in almost all cases. The other baselines present an inconsistent behavior, showing desired performance in certain tasks or for specific $\alpha$ values, and worse performance in other cases. This is attributed to the fact that for the baselines the way the configurations are distributed over the Pareto front is arbitrary. Therefore, sometimes by chance we obtain configurations that are near the testing limit (hence efficient), while in other cases the nearest configuration is far-away (inefficient). On the contrary, the proposed method obtains a dense sampling of the relevant part of the Pareto front, which results in tighter and more stable control across different conditions.}\looseness=-1     

\rev{\newpar{Additional Results} We consider a three-objective scenario of selective classification and robustness, constraining the average error and the miscoverage rate and minimizing the worst error. We see on Figs.~\ref{fig:res_three_obj} and~\ref{fig:res_contorol_three_add} that the proposed method outperforms the baselines. We also explore the budget for which we can match the performance of a dense uniform grid (with over $6$K points) in Fig.~\ref{fig:res_vary_N}. We show that $N=50$ is sufficient, highlighting the computational advantage of the proposed method. In addition, we examine the influence of $\delta'$ in Fig.~\ref{fig:res_delta_prime}, showing that the method is generally insensitive to $\delta'$. Finally, Fig.~\ref{fig:res_sided} shows that using the proposed region is preferable over a single-sided upper bound, implying that it is important to exclude inefficient configurations. \looseness=-1 } 


\section{Conclusion}
We present a flexible framework for reliable model selection that satisfy statistical risk constraints, while optimizing additional conflicting metrics. We define a confined region in the objective space that is a promising target for testing, and propose a \ac{BO} method that identifies Pareto optimal configurations within this region. By statistically validating the candidate set via multiple hypothesis testing, we obtain verified control guarantees. Our experiments have demonstrated the effectiveness of our approach for tuning different types of hyperparameters across various tasks and objectives, including high-accuracy, fairness, robustness, generation and reconstruction quality and cost considerations.\looseness=-1

\bibliography{iclr2024_conference}

\begin{thebibliography}{73}
\providecommand{\natexlab}[1]{#1}
\providecommand{\url}[1]{\texttt{#1}}
\expandafter\ifx\csname urlstyle\endcsname\relax
  \providecommand{\doi}[1]{doi: #1}\else
  \providecommand{\doi}{doi: \begingroup \urlstyle{rm}\Url}\fi

\bibitem[Angelopoulos et~al.(2021)Angelopoulos, Bates, Cand{\`e}s, Jordan, and
  Lei]{angelopoulos2021learn}
Anastasios~N Angelopoulos, Stephen Bates, Emmanuel~J Cand{\`e}s, Michael~I
  Jordan, and Lihua Lei.
\newblock Learn then test: Calibrating predictive algorithms to achieve risk
  control.
\newblock \emph{arXiv preprint arXiv:2110.01052}, 2021.

\bibitem[Angelopoulos et~al.(2022)Angelopoulos, Bates, Fisch, Lei, and
  Schuster]{angelopoulos2022conformal}
Anastasios~N Angelopoulos, Stephen Bates, Adam Fisch, Lihua Lei, and Tal
  Schuster.
\newblock Conformal risk control.
\newblock \emph{arXiv preprint arXiv:2208.02814}, 2022.

\bibitem[Auer(2002)]{auer2002using}
Peter Auer.
\newblock Using confidence bounds for exploitation-exploration trade-offs.
\newblock \emph{Journal of Machine Learning Research}, 3\penalty0
  (Nov):\penalty0 397--422, 2002.

\bibitem[Bader \& Zitzler(2011)Bader and Zitzler]{bader2011hype}
Johannes Bader and Eckart Zitzler.
\newblock Hype: An algorithm for fast hypervolume-based many-objective
  optimization.
\newblock \emph{Evolutionary computation}, 19\penalty0 (1):\penalty0 45--76,
  2011.

\bibitem[Barber et~al.(2021)Barber, Candes, Ramdas, and
  Tibshirani]{barber2021predictive}
Rina~Foygel Barber, Emmanuel~J Candes, Aaditya Ramdas, and Ryan~J Tibshirani.
\newblock Predictive inference with the jackknife+.
\newblock \emph{The Annals of Statistics}, 49\penalty0 (1):\penalty0 486--507,
  2021.

\bibitem[Bates et~al.(2021)Bates, Angelopoulos, Lei, Malik, and
  Jordan]{bates2021distribution}
Stephen Bates, Anastasios Angelopoulos, Lihua Lei, Jitendra Malik, and Michael
  Jordan.
\newblock Distribution-free, risk-controlling prediction sets.
\newblock \emph{Journal of the ACM (JACM)}, 68\penalty0 (6):\penalty0 1--34,
  2021.

\bibitem[Belakaria et~al.(2019)Belakaria, Deshwal, and Doppa]{belakaria2019max}
Syrine Belakaria, Aryan Deshwal, and Janardhan~Rao Doppa.
\newblock Max-value entropy search for multi-objective bayesian optimization.
\newblock In \emph{Advances in Neural Information Processing Systems},
  volume~32, 2019.

\bibitem[Belakaria et~al.(2020)Belakaria, Deshwal, Jayakodi, and
  Doppa]{belakaria2020uncertainty}
Syrine Belakaria, Aryan Deshwal, Nitthilan~Kannappan Jayakodi, and
  Janardhan~Rao Doppa.
\newblock Uncertainty-aware search framework for multi-objective bayesian
  optimization.
\newblock \emph{Proceedings of the AAAI Conference on Artificial Intelligence},
  34\penalty0 (06):\penalty0 10044--10052, 2020.

\bibitem[Bischl et~al.(2023)Bischl, Binder, Lang, Pielok, Richter, Coors,
  Thomas, Ullmann, Becker, Boulesteix, et~al.]{bischl2023hyperparameter}
Bernd Bischl, Martin Binder, Michel Lang, Tobias Pielok, Jakob Richter, Stefan
  Coors, Janek Thomas, Theresa Ullmann, Marc Becker, Anne-Laure Boulesteix,
  et~al.
\newblock Hyperparameter optimization: Foundations, algorithms, best practices,
  and open challenges.
\newblock \emph{Wiley Interdisciplinary Reviews: Data Mining and Knowledge
  Discovery}, 13\penalty0 (2):\penalty0 e1484, 2023.

\bibitem[Chadebec et~al.(2022)Chadebec, Vincent, and
  Allassonniere]{chadebec2022pythae}
Cl\'{e}ment Chadebec, Louis Vincent, and Stephanie Allassonniere.
\newblock Pythae: Unifying generative autoencoders in python - a benchmarking
  use case.
\newblock In S.~Koyejo, S.~Mohamed, A.~Agarwal, D.~Belgrave, K.~Cho, and A.~Oh
  (eds.), \emph{Advances in Neural Information Processing Systems}, volume~35,
  pp.\  21575--21589. Curran Associates, Inc., 2022.

\bibitem[Chen \& Kwok(2022)Chen and Kwok]{chen2022multi}
Weiyu Chen and James Kwok.
\newblock Multi-objective deep learning with adaptive reference vectors.
\newblock \emph{Advances in Neural Information Processing Systems},
  35:\penalty0 32723--32735, 2022.

\bibitem[Cristescu \& Knowles(2015)Cristescu and
  Knowles]{cristescu2015surrogate}
Cristina Cristescu and Joshua Knowles.
\newblock Surrogate-based multiobjective optimization: Parego update and test.
\newblock In \emph{Workshop on Computational Intelligence (UKCI)}, volume 770,
  2015.

\bibitem[Daulton et~al.(2021)Daulton, Balandat, and
  Bakshy]{daulton2021parallel}
Samuel Daulton, Maximilian Balandat, and Eytan Bakshy.
\newblock Parallel bayesian optimization of multiple noisy objectives with
  expected hypervolume improvement.
\newblock \emph{Advances in Neural Information Processing Systems},
  34:\penalty0 2187--2200, 2021.

\bibitem[Deb et~al.(2002)Deb, Pratap, Agarwal, and Meyarivan]{deb2002fast}
Kalyanmoy Deb, Amrit Pratap, Sameer Agarwal, and TAMT Meyarivan.
\newblock A fast and elitist multiobjective genetic algorithm: Nsga-ii.
\newblock \emph{IEEE transactions on evolutionary computation}, 6\penalty0
  (2):\penalty0 182--197, 2002.

\bibitem[D{\'e}sid{\'e}ri(2012)]{desideri2012multiple}
Jean-Antoine D{\'e}sid{\'e}ri.
\newblock Multiple-gradient descent algorithm (mgda) for multiobjective
  optimization.
\newblock \emph{Comptes Rendus Mathematique}, 350\penalty0 (5-6):\penalty0
  313--318, 2012.

\bibitem[Devlin et~al.(2018)Devlin, Chang, Lee, and Toutanova]{devlin2018bert}
Jacob Devlin, Ming-Wei Chang, Kenton Lee, and Kristina Toutanova.
\newblock Bert: Pre-training of deep bidirectional transformers for language
  understanding.
\newblock \emph{arXiv preprint arXiv:1810.04805}, 2018.

\bibitem[Dua et~al.(2017)Dua, Graff, et~al.]{dua2017uci}
Dheeru Dua, Casey Graff, et~al.
\newblock Uci machine learning repository, 2017.

\bibitem[Emmerich et~al.(2005)Emmerich, Beume, and Naujoks]{emmerich2005emo}
Michael Emmerich, Nicola Beume, and Boris Naujoks.
\newblock An emo algorithm using the hypervolume measure as selection
  criterion.
\newblock In \emph{International Conference on Evolutionary Multi-Criterion
  Optimization}, pp.\  62--76. Springer, 2005.

\bibitem[Emmerich et~al.(2006)Emmerich, Giannakoglou, and
  Naujoks]{emmerich2006single}
Michael~TM Emmerich, Kyriakos~C Giannakoglou, and Boris Naujoks.
\newblock Single-and multiobjective evolutionary optimization assisted by
  gaussian random field metamodels.
\newblock \emph{IEEE Transactions on Evolutionary Computation}, 10\penalty0
  (4):\penalty0 421--439, 2006.

\bibitem[Feurer et~al.(2023)Feurer, Eggensperger, Bergman, Pfisterer, Bischl,
  and Hutter]{feurer2023mind}
Matthias Feurer, Katharina Eggensperger, Edward Bergman, Florian Pfisterer,
  Bernd Bischl, and Frank Hutter.
\newblock Mind the gap: Measuring generalization performance across multiple
  objectives.
\newblock In \emph{International Symposium on Intelligent Data Analysis}, pp.\
  130--142. Springer, 2023.

\bibitem[Frazier(2018)]{frazier2018tutorial}
Peter~I Frazier.
\newblock A tutorial on bayesian optimization.
\newblock \emph{arXiv preprint arXiv:1807.02811}, 2018.

\bibitem[Gardner et~al.(2014)Gardner, Kusner, Weinberger, Cunningham,
  et~al.]{gardner2014bayesian}
Jacob Gardner, Matt Kusner, Kilian Weinberger, John Cunningham, et~al.
\newblock Bayesian optimization with inequality constraints.
\newblock In \emph{International Conference on Machine Learning}, pp.\
  937--945. PMLR, 2014.

\bibitem[Gupta et~al.(2020)Gupta, Podkopaev, and Ramdas]{gupta2020binary}
Chirag Gupta, Aleksandr Podkopaev, and Aaditya Ramdas.
\newblock Distribution-free binary classification: prediction sets, confidence
  intervals and calibration.
\newblock In \emph{Advances in Neural Information Processing Systems
  (NeurIPS)}, 2020.

\bibitem[Hardt et~al.(2016)Hardt, Price, and Srebro]{hardt2016equality}
Moritz Hardt, Eric Price, and Nati Srebro.
\newblock Equality of opportunity in supervised learning.
\newblock \emph{Advances in neural information processing systems}, 29, 2016.

\bibitem[Hern{\'a}ndez-Lobato et~al.(2016)Hern{\'a}ndez-Lobato,
  Hernandez-Lobato, Shah, and Adams]{hernandez2016predictive}
Daniel Hern{\'a}ndez-Lobato, Jose Hernandez-Lobato, Amar Shah, and Ryan Adams.
\newblock Predictive entropy search for multi-objective bayesian optimization.
\newblock In \emph{International conference on machine learning}, pp.\
  1492--1501. PMLR, 2016.

\bibitem[Higgins et~al.(2016)Higgins, Matthey, Pal, Burgess, Glorot, Botvinick,
  Mohamed, and Lerchner]{higgins2016beta}
Irina Higgins, Loic Matthey, Arka Pal, Christopher Burgess, Xavier Glorot,
  Matthew Botvinick, Shakir Mohamed, and Alexander Lerchner.
\newblock beta-vae: Learning basic visual concepts with a constrained
  variational framework.
\newblock In \emph{International conference on learning representations}, 2016.

\bibitem[Hoeffding(1994)]{hoeffding1994probability}
Wassily Hoeffding.
\newblock Probability inequalities for sums of bounded random variables.
\newblock \emph{The collected works of Wassily Hoeffding}, pp.\  409--426,
  1994.

\bibitem[Holm(1979)]{holm1979simple}
Sture Holm.
\newblock A simple sequentially rejective multiple test procedure.
\newblock \emph{Scandinavian journal of statistics}, pp.\  65--70, 1979.

\bibitem[Ishibuchi et~al.(2018)Ishibuchi, Imada, Setoguchi, and
  Nojima]{ishibuchi2018specify}
Hisao Ishibuchi, Ryo Imada, Yu~Setoguchi, and Yusuke Nojima.
\newblock How to specify a reference point in hypervolume calculation for fair
  performance comparison.
\newblock \emph{Evolutionary computation}, 26\penalty0 (3):\penalty0 411--440,
  2018.

\bibitem[Izmailov et~al.(2022)Izmailov, Kirichenko, Gruver, and
  Wilson]{izmailov2022feature}
Pavel Izmailov, Polina Kirichenko, Nate Gruver, and Andrew~G Wilson.
\newblock On feature learning in the presence of spurious correlations.
\newblock \emph{Advances in Neural Information Processing Systems},
  35:\penalty0 38516--38532, 2022.

\bibitem[Karl et~al.(2022)Karl, Pielok, Moosbauer, Pfisterer, Coors, Binder,
  Schneider, Thomas, Richter, Lang, et~al.]{karl2022multi}
Florian Karl, Tobias Pielok, Julia Moosbauer, Florian Pfisterer, Stefan Coors,
  Martin Binder, Lennart Schneider, Janek Thomas, Jakob Richter, Michel Lang,
  et~al.
\newblock Multi-objective hyperparameter optimization--an overview.
\newblock \emph{arXiv preprint arXiv:2206.07438}, 2022.

\bibitem[Kingma \& Welling(2013)Kingma and Welling]{kingma2013auto}
Diederik~P Kingma and Max Welling.
\newblock Auto-encoding variational bayes.
\newblock \emph{arXiv preprint arXiv:1312.6114}, 2013.

\bibitem[Knowles(2006)]{knowles2006parego}
Joshua Knowles.
\newblock Parego: A hybrid algorithm with on-line landscape approximation for
  expensive multiobjective optimization problems.
\newblock \emph{IEEE Transactions on Evolutionary Computation}, 10\penalty0
  (1):\penalty0 50--66, 2006.

\bibitem[Konakovic~Lukovic et~al.(2020)Konakovic~Lukovic, Tian, and
  Matusik]{konakovic2020diversity}
Mina Konakovic~Lukovic, Yunsheng Tian, and Wojciech Matusik.
\newblock Diversity-guided multi-objective bayesian optimization with batch
  evaluations.
\newblock \emph{Advances in Neural Information Processing Systems},
  33:\penalty0 17708--17720, 2020.

\bibitem[Kushner(1964)]{kushner1964new}
HJ~Kushner.
\newblock A new method of locating the maximum point of an arbitrary multipeak
  curve in the presence of noise.
\newblock \emph{Journal of Basic Engineering}, 86\penalty0 (1):\penalty0
  97--106, 1964.

\bibitem[Laskaridis et~al.(2021)Laskaridis, Kouris, and
  Lane]{laskaridis2021adaptive}
Stefanos Laskaridis, Alexandros Kouris, and Nicholas~D Lane.
\newblock Adaptive inference through early-exit networks: Design, challenges
  and directions.
\newblock In \emph{Proceedings of the 5th International Workshop on Embedded
  and Mobile Deep Learning}, pp.\  1--6, 2021.

\bibitem[Laufer-Goldshtein et~al.(2023)Laufer-Goldshtein, Fisch, Barzilay, and
  Jaakkola]{laufer2023efficiently}
Bracha Laufer-Goldshtein, Adam Fisch, Regina Barzilay, and Tommi Jaakkola.
\newblock Efficiently controlling multiple risks with pareto testing.
\newblock \emph{ICLR}, 2023.

\bibitem[LeCun(1998)]{lecun1998mnist}
Yann LeCun.
\newblock The mnist database of handwritten digits.
\newblock \emph{http://yann. lecun. com/exdb/mnist/}, 1998.

\bibitem[Lei et~al.(2013)Lei, Robins, and Wasserman]{lei-robins-wasserman-dfps}
Jing Lei, James Robins, and Larry Wasserman.
\newblock Distribution-free prediction sets.
\newblock \emph{Journal of the American Statistical Association}, 108\penalty0
  (501):\penalty0 278--287, 2013.

\bibitem[Lei et~al.(2018)Lei, G’Sell, Rinaldo, Tibshirani, and
  Wasserman]{lei2018distribution}
Jing Lei, Max G’Sell, Alessandro Rinaldo, Ryan~J Tibshirani, and Larry
  Wasserman.
\newblock Distribution-free predictive inference for regression.
\newblock \emph{Journal of the American Statistical Association}, 113\penalty0
  (523):\penalty0 1094--1111, 2018.

\bibitem[Letham et~al.(2019)Letham, Karrer, Ottoni, and
  Bakshy]{letham2019constrained}
Benjamin Letham, Brian Karrer, Guilherme Ottoni, and Eytan Bakshy.
\newblock Constrained bayesian optimization with noisy experiments.
\newblock \emph{Bayesian Analysis}, 14\penalty0 (2), 2019.

\bibitem[Lin et~al.(2019)Lin, Zhen, Li, Zhang, and Kwong]{lin2019pareto}
Xi~Lin, Hui-Ling Zhen, Zhenhua Li, Qing-Fu Zhang, and Sam Kwong.
\newblock Pareto multi-task learning.
\newblock \emph{Advances in neural information processing systems}, 32, 2019.

\bibitem[Lin et~al.(2020)Lin, Yang, Zhang, and Kwong]{lin2020controllable}
Xi~Lin, Zhiyuan Yang, Qingfu Zhang, and Sam Kwong.
\newblock Controllable pareto multi-task learning.
\newblock \emph{arXiv preprint arXiv:2010.06313}, 2020.

\bibitem[Lindauer et~al.(2022)Lindauer, Eggensperger, Feurer, Biedenkapp, Deng,
  Benjamins, Ruhkopf, Sass, and Hutter]{lindauer2022smac3}
Marius Lindauer, Katharina Eggensperger, Matthias Feurer, Andr{\'e} Biedenkapp,
  Difan Deng, Carolin Benjamins, Tim Ruhkopf, Ren{\'e} Sass, and Frank Hutter.
\newblock Smac3: A versatile bayesian optimization package for hyperparameter
  optimization.
\newblock \emph{J. Mach. Learn. Res.}, 23:\penalty0 54--1, 2022.

\bibitem[Lohaus et~al.(2020)Lohaus, Perrot, and Von~Luxburg]{lohaus2020too}
Michael Lohaus, Michael Perrot, and Ulrike Von~Luxburg.
\newblock Too relaxed to be fair.
\newblock In \emph{International Conference on Machine Learning}, pp.\
  6360--6369. PMLR, 2020.

\bibitem[Mahapatra \& Rajan(2020)Mahapatra and Rajan]{mahapatra2020multi}
Debabrata Mahapatra and Vaibhav Rajan.
\newblock Multi-task learning with user preferences: Gradient descent with
  controlled ascent in pareto optimization.
\newblock In \emph{International Conference on Machine Learning}, pp.\
  6597--6607. PMLR, 2020.

\bibitem[McKay et~al.(2000)McKay, Beckman, and Conover]{mckay2000comparison}
Michael~D McKay, Richard~J Beckman, and William~J Conover.
\newblock A comparison of three methods for selecting values of input variables
  in the analysis of output from a computer code.
\newblock \emph{Technometrics}, 42\penalty0 (1):\penalty0 55--61, 2000.

\bibitem[Menghani(2023)]{menghani2023efficient}
Gaurav Menghani.
\newblock Efficient deep learning: A survey on making deep learning models
  smaller, faster, and better.
\newblock \emph{ACM Computing Surveys}, 55\penalty0 (12):\penalty0 1--37, 2023.

\bibitem[Mo{\v{c}}kus(1975)]{movckus1975bayesian}
Jonas Mo{\v{c}}kus.
\newblock On bayesian methods for seeking the extremum.
\newblock In \emph{Optimization Techniques IFIP Technical Conference:
  Novosibirsk, July 1--7, 1974}, pp.\  400--404. Springer, 1975.

\bibitem[Navon et~al.(2020)Navon, Shamsian, Chechik, and
  Fetaya]{navon2020learning}
Aviv Navon, Aviv Shamsian, Gal Chechik, and Ethan Fetaya.
\newblock Learning the pareto front with hypernetworks.
\newblock \emph{arXiv preprint arXiv:2010.04104}, 2020.

\bibitem[Padh et~al.(2021)Padh, Antognini, Lejal-Glaude, Faltings, and
  Musat]{padh2021addressing}
Kirtan Padh, Diego Antognini, Emma Lejal-Glaude, Boi Faltings, and Claudiu
  Musat.
\newblock Addressing fairness in classification with a model-agnostic
  multi-objective algorithm.
\newblock In \emph{Uncertainty in Artificial Intelligence}, pp.\  600--609.
  PMLR, 2021.

\bibitem[Paria et~al.(2020)Paria, Kandasamy, and P{\'o}czos]{paria2020flexible}
Biswajit Paria, Kirthevasan Kandasamy, and Barnab{\'a}s P{\'o}czos.
\newblock A flexible framework for multi-objective bayesian optimization using
  random scalarizations.
\newblock In \emph{Uncertainty in Artificial Intelligence}, pp.\  766--776.
  PMLR, 2020.

\bibitem[Pessach \& Shmueli(2022)Pessach and Shmueli]{pessach2022review}
Dana Pessach and Erez Shmueli.
\newblock A review on fairness in machine learning.
\newblock \emph{ACM Computing Surveys (CSUR)}, 55\penalty0 (3):\penalty0 1--44,
  2022.

\bibitem[Ponweiser et~al.(2008)Ponweiser, Wagner, Biermann, and
  Vincze]{ponweiser2008multiobjective}
Wolfgang Ponweiser, Tobias Wagner, Dirk Biermann, and Markus Vincze.
\newblock Multiobjective optimization on a limited budget of evaluations using
  model-assisted-metric selection.
\newblock In \emph{International conference on parallel problem solving from
  nature}, pp.\  784--794. Springer, 2008.

\bibitem[Rezende et~al.(2014)Rezende, Mohamed, and
  Wierstra]{rezende2014stochastic}
Danilo~Jimenez Rezende, Shakir Mohamed, and Daan Wierstra.
\newblock Stochastic backpropagation and approximate inference in deep
  generative models.
\newblock In \emph{International conference on machine learning}, pp.\
  1278--1286. PMLR, 2014.

\bibitem[Ruchte \& Grabocka(2021)Ruchte and Grabocka]{ruchte2021scalable}
Michael Ruchte and Josif Grabocka.
\newblock Scalable pareto front approximation for deep multi-objective
  learning.
\newblock In \emph{2021 IEEE international conference on data mining (ICDM)},
  pp.\  1306--1311. IEEE, 2021.

\bibitem[Sagawa et~al.(2019)Sagawa, Koh, Hashimoto, and
  Liang]{sagawa2019distributionally}
Shiori Sagawa, Pang~Wei Koh, Tatsunori~B Hashimoto, and Percy Liang.
\newblock Distributionally robust neural networks for group shifts: On the
  importance of regularization for worst-case generalization.
\newblock \emph{arXiv preprint arXiv:1911.08731}, 2019.

\bibitem[Salinas et~al.(2023)Salinas, Golebiowski, Klein, Seeger, and
  Archambeau]{salinas2023optimizing}
David Salinas, Jacek Golebiowski, Aaron Klein, Matthias Seeger, and Cedric
  Archambeau.
\newblock Optimizing hyperparameters with conformal quantile regression.
\newblock \emph{arXiv preprint arXiv:2305.03623}, 2023.

\bibitem[Sener \& Koltun(2018)Sener and Koltun]{sener2018multi}
Ozan Sener and Vladlen Koltun.
\newblock Multi-task learning as multi-objective optimization.
\newblock \emph{Advances in neural information processing systems}, 31, 2018.

\bibitem[Shahriari et~al.(2015)Shahriari, Swersky, Wang, Adams, and
  De~Freitas]{shahriari2015taking}
Bobak Shahriari, Kevin Swersky, Ziyu Wang, Ryan~P Adams, and Nando De~Freitas.
\newblock Taking the human out of the loop: A review of bayesian optimization.
\newblock \emph{Proceedings of the IEEE}, 104\penalty0 (1):\penalty0 148--175,
  2015.

\bibitem[Stanton et~al.(2023)Stanton, Maddox, and Wilson]{stanton2023bayesian}
S~Stanton, W~Maddox, and AG~Wilson.
\newblock Bayesian optimization with conformal prediction sets.
\newblock In \emph{Artificial Intelligence and Statistics}, 2023.

\bibitem[Vovk(2002)]{vovk2002calibration}
Vladimir Vovk.
\newblock On-line confidence machines are well-calibrated.
\newblock In \emph{The 43rd Annual IEEE Symposium on Foundations of Computer
  Science.}, 2002.

\bibitem[Vovk et~al.(2015)Vovk, Petej, and Fedorova]{vovk2015probabilistic}
Vladimir Vovk, Ivan Petej, and Valentina Fedorova.
\newblock Large-scale probabilistic predictors with and without guarantees of
  validity.
\newblock In \emph{Advances in Neural Information Processing Systems
  (NeurIPS)}, 2015.

\bibitem[Vovk et~al.(2017)Vovk, Shen, Manokhin, and Xie]{pmlr-v60-vovk17a}
Vladimir Vovk, Jieli Shen, Valery Manokhin, and Min-ge Xie.
\newblock Nonparametric predictive distributions based on conformal prediction.
\newblock In \emph{Proceedings of the Sixth Workshop on Conformal and
  Probabilistic Prediction and Applications}, 2017.

\bibitem[Wang et~al.(2022)Wang, Jin, Schmitt, and Olhofer]{wang2022recent}
Xilu Wang, Yaochu Jin, Sebastian Schmitt, and Markus Olhofer.
\newblock Recent advances in bayesian optimization.
\newblock \emph{arXiv preprint arXiv:2206.03301}, 2022.

\bibitem[Williams \& Rasmussen(2006)Williams and
  Rasmussen]{williams2006gaussian}
Christopher~KI Williams and Carl~Edward Rasmussen.
\newblock \emph{Gaussian processes for machine learning}.
\newblock MIT press Cambridge, MA, 2006.

\bibitem[Wo{\l}czyk et~al.(2021)Wo{\l}czyk, W{\'o}jcik, Ba{\l}azy, Podolak,
  Tabor, {\'S}mieja, and Trzcinski]{wolczyk2021zero}
Maciej Wo{\l}czyk, Bartosz W{\'o}jcik, Klaudia Ba{\l}azy, Igor~T Podolak, Jacek
  Tabor, Marek {\'S}mieja, and Tomasz Trzcinski.
\newblock Zero time waste: Recycling predictions in early exit neural networks.
\newblock \emph{Advances in Neural Information Processing Systems},
  34:\penalty0 2516--2528, 2021.

\bibitem[Wu et~al.(2019)Wu, Zhang, and Wu]{wu2019convexity}
Yongkai Wu, Lu~Zhang, and Xintao Wu.
\newblock On convexity and bounds of fairness-aware classification.
\newblock In \emph{The World Wide Web Conference}, pp.\  3356--3362, 2019.

\bibitem[Yang et~al.(2023)Yang, Zhang, Katabi, and Ghassemi]{yang2023change}
Yuzhe Yang, Haoran Zhang, Dina Katabi, and Marzyeh Ghassemi.
\newblock Change is hard: A closer look at subpopulation shift.
\newblock \emph{arXiv preprint arXiv:2302.12254}, 2023.

\bibitem[Zhang \& Li(2007)Zhang and Li]{zhang2007moea}
Qingfu Zhang and Hui Li.
\newblock Moea/d: A multiobjective evolutionary algorithm based on
  decomposition.
\newblock \emph{IEEE Transactions on evolutionary computation}, 11\penalty0
  (6):\penalty0 712--731, 2007.

\bibitem[Zhang et~al.(2015)Zhang, Zhao, and LeCun]{zhang2015character}
Xiang Zhang, Junbo Zhao, and Yann LeCun.
\newblock Character-level convolutional networks for text classification.
\newblock \emph{Advances in neural information processing systems}, 28, 2015.

\bibitem[Zhang et~al.(2023)Zhang, Park, and Simeone]{zhang2023bayesian}
Yunchuan Zhang, Sangwoo Park, and Osvaldo Simeone.
\newblock Bayesian optimization with formal safety guarantees via online
  conformal prediction.
\newblock \emph{arXiv preprint arXiv:2306.17815}, 2023.

\bibitem[Zitzler \& Thiele(1998)Zitzler and Thiele]{zitzler1998multiobjective}
Eckart Zitzler and Lothar Thiele.
\newblock Multiobjective optimization using evolutionary algorithms—a
  comparative case study.
\newblock In \emph{International conference on parallel problem solving from
  nature}, pp.\  292--301. Springer, 1998.

\end{thebibliography}
\bibliographystyle{iclr2024_conference}

\appendix
\counterwithin{figure}{section}
\counterwithin{table}{section}
\counterwithin{algorithm}{section}
\section{Additional related work}
\label{sec:additional_related}
\newpar{Gradient-Based MOO} When dealing with differentiable objective functions, gradient-based \ac{MOO} algorithms can be utilized. The cornerstone of these methods is Multiple-Gradient Descent (MGD)~\cite{sener2018multi, desideri2012multiple}, which ensures that all objectives are decreased simultaneously, leading to convergence at a Pareto optimal point. Several extensions were proposed to enable convergence to a specific point on the front defined by a preference vector ~\cite{lin2019pareto,mahapatra2020multi}, or learning the entire Pareto front, using a preference-conditioned model ~\cite{navon2020learning,lin2020controllable,chen2022multi,ruchte2021scalable}. However, this line of research focuses on differentiable objectives, optimizing the loss space used during training, which is typically different from the ultimate non-differentiable metrics used for evaluation (e.g. error rates). Furthermore, it focuses on recovering a single or multiple (possibly infinitely many) Pareto optimal points, without addressing the actual selection of model configuration under specific constraints, which is the problem we tackle in this paper.\looseness=-1

\section{Implementation and dataset details}
\label{sec:exp_details}
We provide here further details on the datasets, model architectures, training procedures, and examined scenarios, which were used in our experiments. 

\textbf{Datasets and Evaluation Details.} Detailed information on the datasets and the examined scenarios is provided in Table~\ref{tab:dataset}, including: the number of samples for each data split (train/validation/calibration/test), the best and the worst performance for each objective (over the validation set) and the optimization budget. \rev{We emphasize again the purpose of each data split. The training dataset is used for learning the model's parameters. The validation data is used for \ac{BO} in Algorithm~\ref{alg:bo} and for ordering the chosen configurations. The calibration data is used for the testing procedure. The final chosen $\bm{\lambda}^*$~\eqref{eq:select} is used for setting the model configuration. Finally, the performance of the selected model is examined over the test dataset. We repeat the experiments, with different splits to calibration and test sets, and report the scores of the test data, averaged over different trials. We define the range of $\alpha$ bounds according to the values observed for the initial pool of configurations that is generated at the beginning of the \ac{BO} procedure. The minimum and the maximum edge points obtained for each task appear in table~\ref{tab:dataset}, and are also observed in the examples shown on Fig.~\ref{fig:bbo_configs}. We choose values in between these extreme edge points, but not too close to either side, since too small values may not be statistically achievable and too large values are trivially satisfied (with tighter control not significantly improving the free objective).}

\textbf{Fairness.} Our model is a $3$-layer feed-forward neural network with hidden dimensions $[60,25]$. We train all models using Adam optimizer with learning rate $1e-3$ for $50$ epochs and batch size $256$.

\textbf{Robustness.} We use a ResNet-50 model pretrained on ImageNet. We train the models for $50$ epochs with SGD with a constant learning rate of $1e-3$, momentum decay of $0.9$, batch size $32$ and weight decay of $1e-4$. We use random crops and horizontal flips as data augmentation. We use half of the CelebA validation data to train the last layer, and the other half for \ac{BO}.

\textbf{VAE.} We use the implementation provided by~\cite{chadebec2022pythae} of a ResNet-based encoder and decoder, trained using AdamW optimizer with $\beta_1 = 0.91, \beta_2 = 0.99$, and weight decay $0.05$. We set the learning to $1e-4$ and the batch size to $64$. The training process consisted of $10$ epochs. We use binary-cross entropy reconstruction loss for training the model, and the mean squared error normalized by the total number of pixels (728) as the reconstruction objective function for hyperparameter tuning. 

\textbf{Pruning.} We use a BERT-base model~\cite{devlin2018bert} with $12$ layers
and $12$ heads per layer. We follow the recipe in~\cite{laufer2023efficiently} and attach a prediction head and a token importance predictor per layer. The core model is first finetuned on the task. We compute the attention head importance scores based on $5$K held-out samples out of the training data. We freeze the backbone model and train the early-exit classifiers and the token importance predictors on the training data ($115$K samples). 

Each prediction head is a $2$-layer feed-forward neural network with $32$ dimensional hidden states, and ReLU activation. The input is the hidden representation of the \texttt{[CLS]} token concatenated with the hidden representation of all previous layers, following~\citep{wolczyk2021zero}. 

Similarly, each token importance predictor is a $2$-layer feed-forward neural network with $32$ dimensional hidden states, and ReLU activation. The input is the hidden representation of each token in the current layer and all previous layers~\citep{wolczyk2021zero}.


\begin{table}[t]
\caption{Datasets Details}
\label{tab:dataset}
\begin{center}
\resizebox{0.99 \linewidth}{!}{%
\begin{tabular}{c c c c c c c c c c c}
\toprule
\toprule
\multicolumn{1}{c}{Dataset}  & \multicolumn{1}{c}{Train} & \multicolumn{1}{c}{Validation} & \multicolumn{1}{c}{Calibration} & \multicolumn{1}{c}{Test} & \multicolumn{1}{c}{Objectives $(\red{\ell_1},\blue{\ell_2})$}  &\multicolumn{1}{c}{(\red{best $\ell_1$}, \blue{worst $\ell_2$})}  &\multicolumn{1}{c}{(\red{worst $\ell_1$}, \blue{best $\ell_2$})}
&\multicolumn{1}{c}{$N$}&\multicolumn{1}{c}{$N_0$}\\ \hline \\
Adult       & $32{,}559$ & $3{,}618$ & $4{,}522$ & $4{,}523$ & (\red{Error}, \blue{DDP}) & (\red{0.154}, \blue{0.145}) & (\red{0.225}, \blue{0.01})  & $10$ & $5$ &\\
CelebA      & $162{,}770$ & $19{,}867
$ & $9{,}981$ & $9{,}981$ & (\red{Avg. Error}, \blue{Worst Error}) & (\red{0.045}, \blue{0.62}) & (\red{0.089}, \blue{0.11}) & $15$ & $5$&\\
MNIST       &$50{,}000$ & $10{,}000$ & $5{,}000$ & $5{,}000$ &  (\red{Recon. Error}, \blue{KLD}) & (\red{0.008}, \blue{80}) & (\red{0.072}, \blue{0.01}) & $10$ & $5$ &\\
AG News     & $120{,}000$ & $2{,}500$ & $2{,}500$ & $2{,}600$ &  (\red{Acc. Difference}, \blue{Rel. Cost}) & (\red{0.0}, \blue{1.0}) & (\red{0.8}, \blue{0.0}) & $50$ & $30$ &\\
\bottomrule
\bottomrule
\end{tabular}}
\end{center}
\end{table}
\section{Additional Results}
\label{sec:more_res}

\begin{figure}[t]
\begin{center}
\begin{subfigure}[b]{0.45\textwidth}
 \centering
 \includegraphics[width=\textwidth]{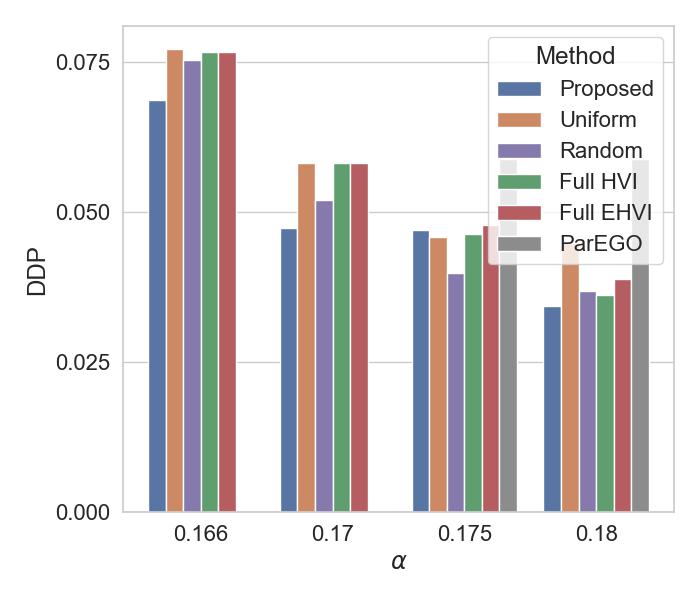}
 \caption{Fairness}
\end{subfigure}
\hspace{-0.9em}
\begin{subfigure}[b]{0.45\textwidth}
 \centering
 \includegraphics[width=\textwidth]{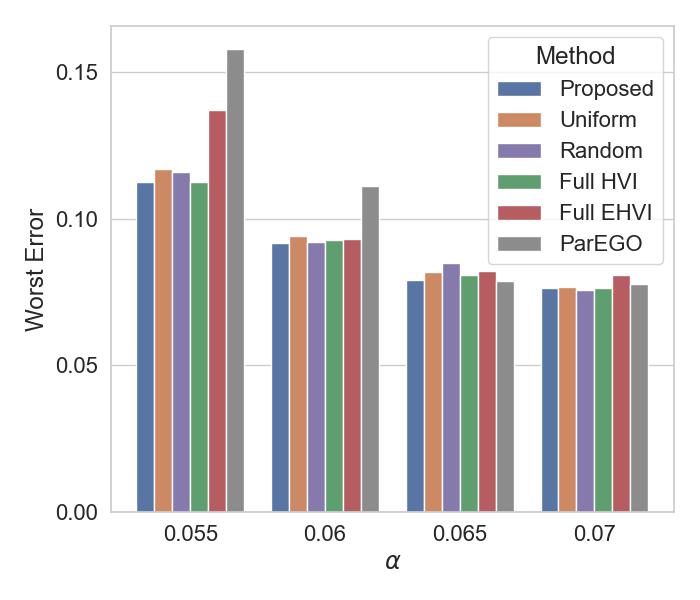}
 \caption{Robustness}
\end{subfigure}
\hspace{-0.9em}
\begin{subfigure}[b]{0.45\textwidth}
 \centering
 \includegraphics[width=\textwidth]{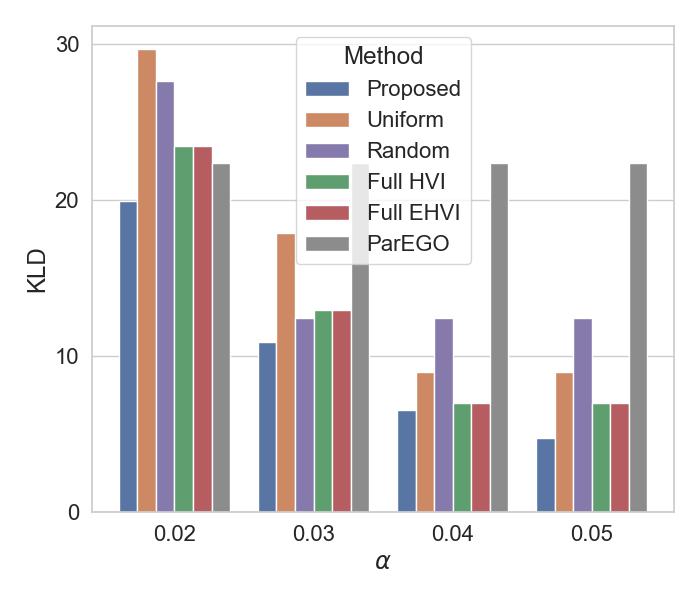}
 \caption{VAE}
 \end{subfigure}
 \hspace{-0.9em}
 \begin{subfigure}[b]{0.45\textwidth}
 \centering
 \includegraphics[width=\textwidth]{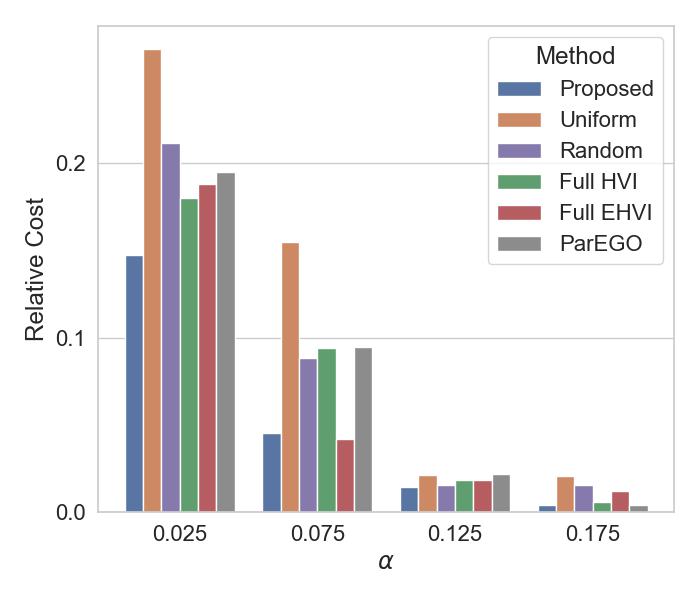}
\caption{Pruning}
\end{subfigure}
\end{center}
\caption{\rev{Two objectives, additional baselines (50 random splits) - presenting the scores obtained for the free objectives. For the fairness task, ParEGO failed to find a valid configurations in $14\%$ and $2\%$ of the runs for $\alpha=0.166$ and $\alpha=0.17$, respectively. Therefore we do not show the scores of to this baseline in these two cases.}}  
\label{fig:res_contorol_two_add}
\end{figure}

\begin{figure}[t!]
\begin{center}
\begin{subfigure}[b]{0.33\textwidth}
 \includegraphics[width=\textwidth]{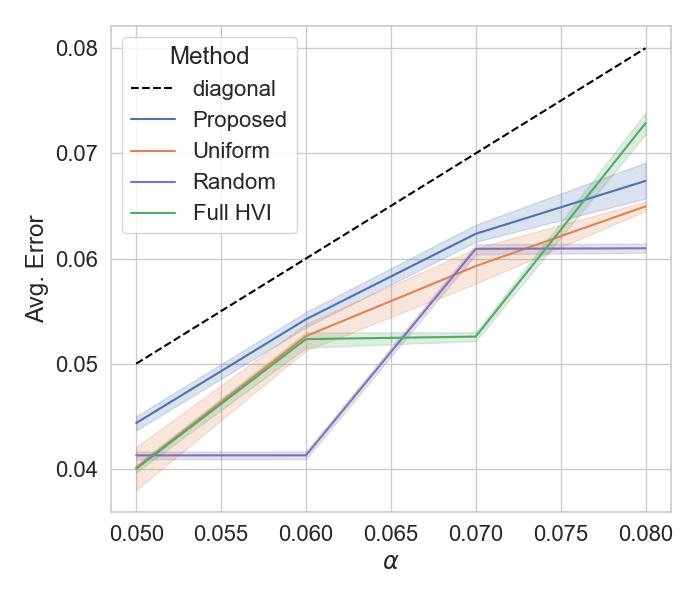}
\end{subfigure}
\hspace{-0.5em}
\begin{subfigure}[b]{0.33\textwidth}
 \includegraphics[width=\textwidth]{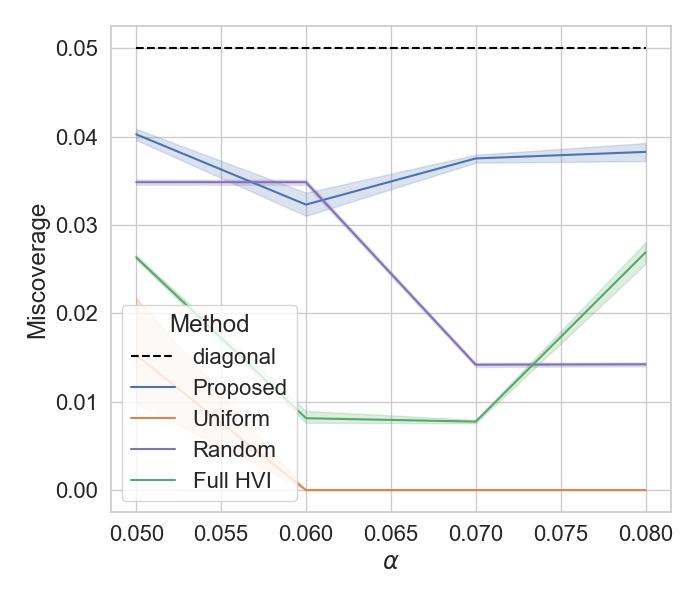}
\end{subfigure}
\hspace{-0.5em}
\begin{subfigure}[b]{0.33\textwidth} \includegraphics[width=\textwidth]{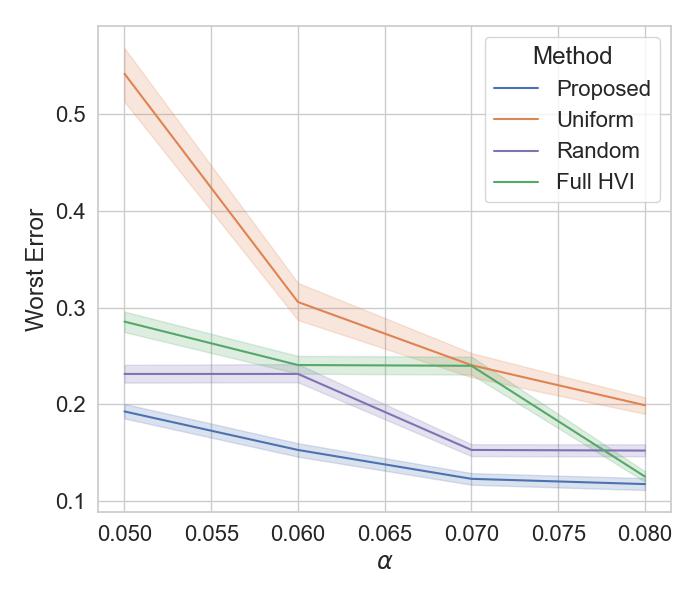}
\end{subfigure}
\end{center}
\vspace{-12pt}
\caption{\rev{Three-objectives, in the task of robustness with selective classification ($50$ random splits): average error limited is by $\alpha_1\in[0.05, 0.08]$, miscoverage limited by $\alpha_2=0.05$, and worst accuracy minimized.\looseness=-1}} 
\label{fig:res_three_obj}
\vspace{-20pt}
\end{figure}

\begin{figure}[t]
\begin{center}
\begin{subfigure}[b]{0.4\textwidth}
 \centering
 \includegraphics[width=\textwidth]{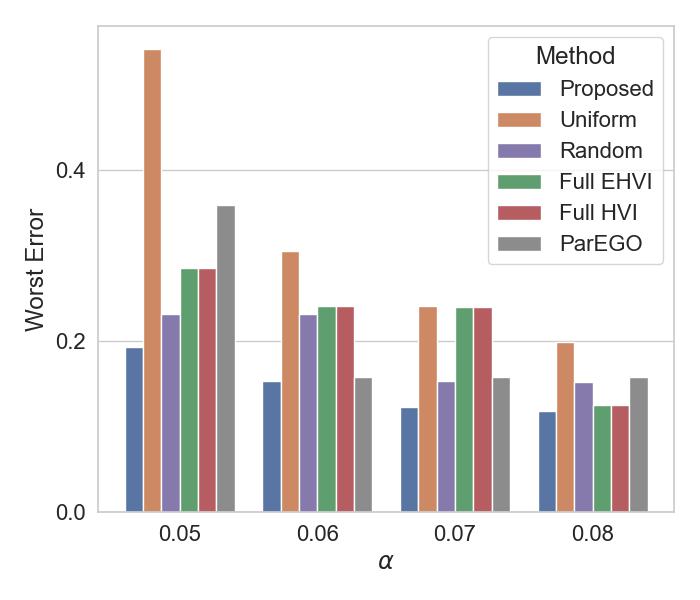}
\end{subfigure}
\end{center}
\caption{\rev{Three objectives, additional baselines - presenting the scores obtained for the free objective - worst accuracy.}}  
\label{fig:res_contorol_three_add}
\end{figure}

In this section, we describe additional experiments and results. 

\begin{figure}[t]
\centering
\begin{subfigure}[b]{0.45\textwidth}
\includegraphics[width=\textwidth]{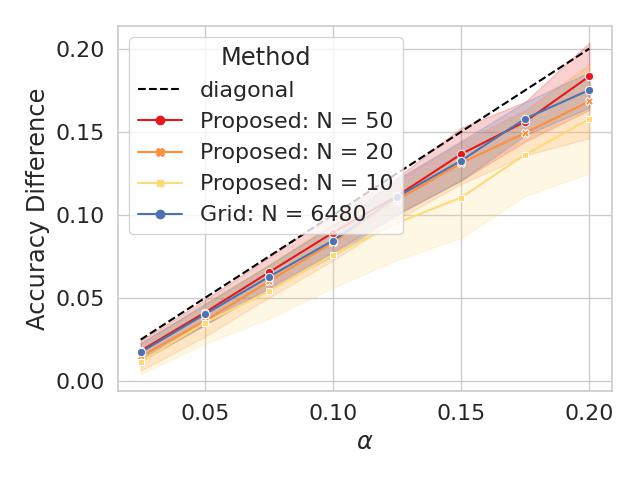}
\end{subfigure}
\begin{subfigure}[b]{0.45\textwidth}
\includegraphics[width=\textwidth]{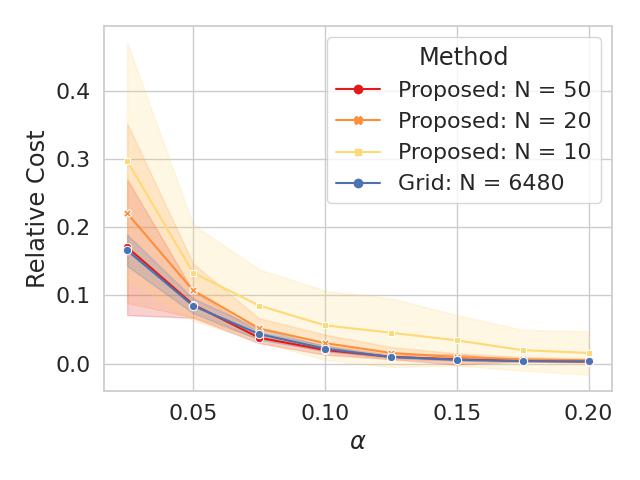}
\end{subfigure}
\caption{Results of the proposed method over AG News for different number of evaluations, and with a grid of thresholds. Results are averaged over $50$ random splits. Accuracy reduction is controlled and cost is minimized. \looseness=-1}
\label{fig:res_vary_N}
\end{figure}

\newpar{Varying Optimization Budget} We examine the effect of varying the optimization budget $N$. We show results for the pruning task with $N\in\{10,20,50\}$. In addition, we compare to a dense grid with uniform sampling of all 3 hyperparmeters with a total of $N=6480$ configurations. We see on Fig.~\ref{fig:res_vary_N} that the relative cost gradually improves with the increase in $N$ and reaches the performance of the dense grid approach with $N=50$. This indicates that using our proposed method we can significantly decrease the required budget without scarifying performance.  

\begin{figure}[t]
\begin{center}
\begin{subfigure}[b]{0.45\textwidth}
 \centering
 \includegraphics[width=\textwidth]{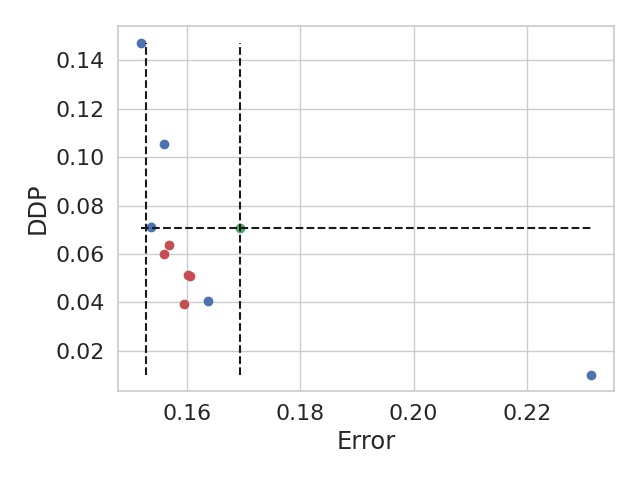}
 \caption{Fairness ($\alpha=0.165$)}
\end{subfigure}
\hspace{-0.9em}
\begin{subfigure}[b]{0.45\textwidth}
 \centering
 \includegraphics[width=\textwidth]{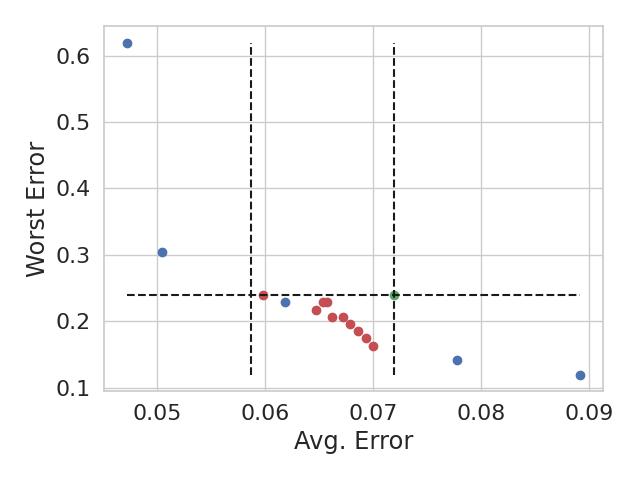}
 \caption{Robustness ($\alpha=0.07$)}
\end{subfigure}
\hspace{-0.9em}
\begin{subfigure}[b]{0.45\textwidth}
 \centering
 \includegraphics[width=\textwidth]{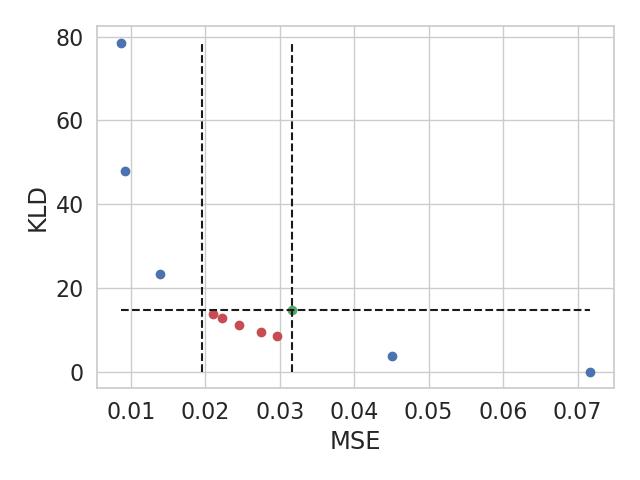}
 \caption{VAE ($\alpha=0.03$)}
 \end{subfigure}
 \hspace{-0.9em}
 \begin{subfigure}[b]{0.45\textwidth}
 \centering
 \includegraphics[width=\textwidth]{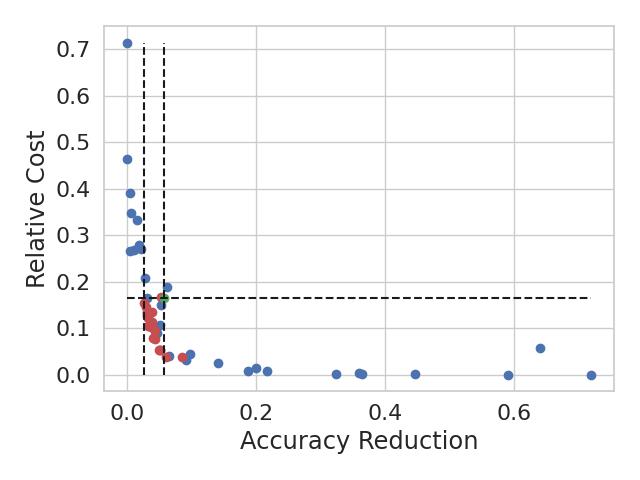}
\caption{Pruning ($\alpha=0.05$)}
\end{subfigure}
\end{center}
\caption{Demonstration of the selection outcomes of the proposed \ac{BO}: the green point is the defined reference point, the blue points correspond to the initial set of configurations, and the red points correspond to selected configurations. Dashed lines enclose the region of interest.}  
\label{fig:bbo_configs}
\end{figure}

\newpar{Demonstration of BO Selection} We show the outcomes of the proposed \ac{BO} procedure across different tasks in Fig.~\ref{fig:bbo_configs}. The reference point defined in~\eqref{eq:ref_point} is marked in green, and the boundaries of the region of interest are depicted by the dashed lines. The blue points correspond to the configurations in the initial pool $\mathcal{C}_0$, while the red points correspond to the configurations selected by our \ac{BO} procedure. We see that the specified region is significantly smaller compared to the entire front. Moreover, we observe that through \ac{BO} we obtain a dense set of configurations that is confined to the region of interest as desired. 

\begin{figure}[t]
\begin{center}
\begin{subfigure}[b]{0.25\textwidth}
 \includegraphics[width=\textwidth]{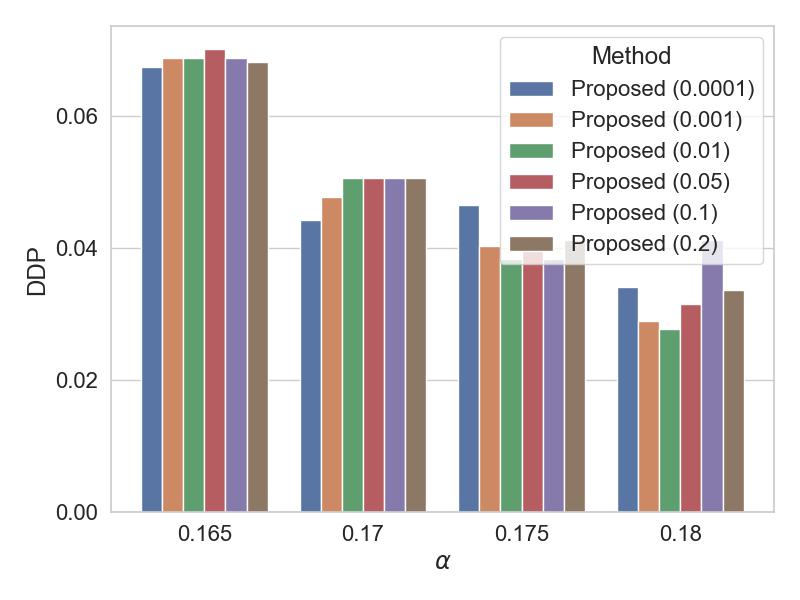}
\caption{Fairness}
\end{subfigure}
\hspace{-0.5em}
\begin{subfigure}[b]{0.25\textwidth}
 \includegraphics[width=\textwidth]{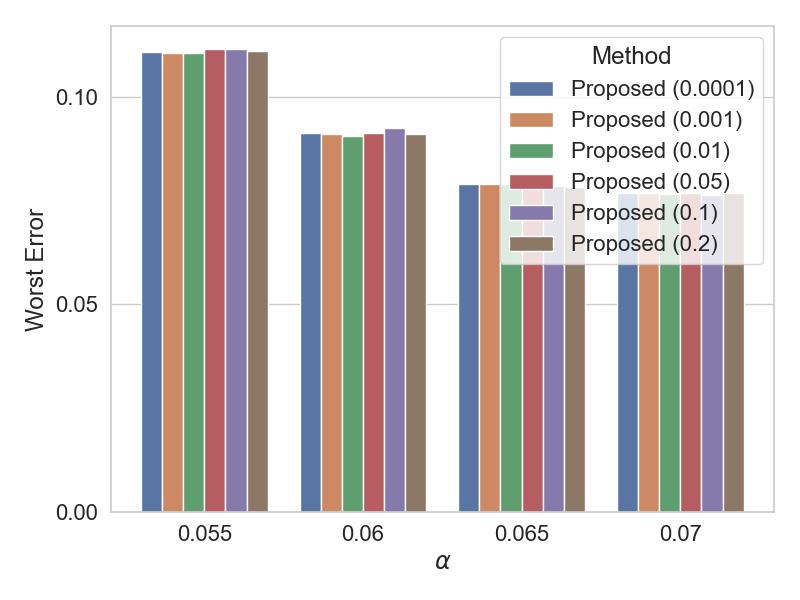}
\caption{Robustness}
\end{subfigure}
\hspace{-0.5em}
\begin{subfigure}[b]{0.25\textwidth}
 \includegraphics[width=\textwidth]{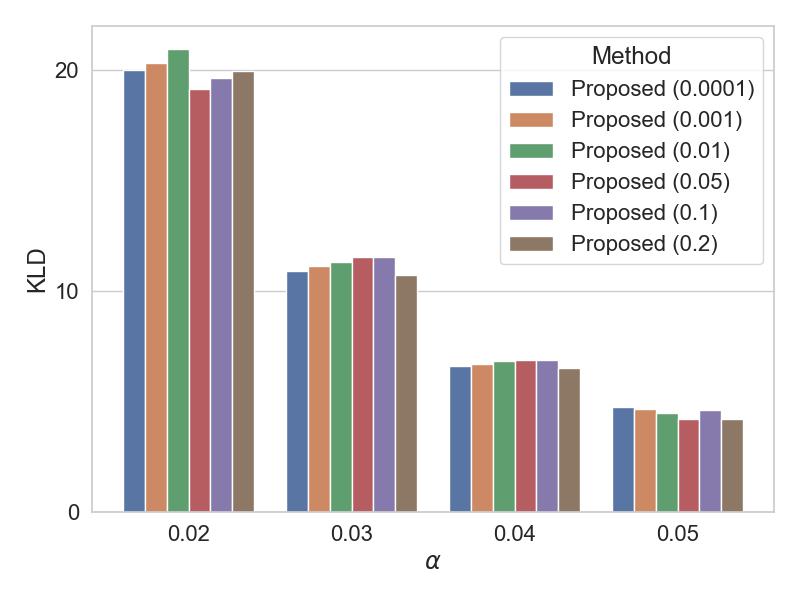}
\caption{VAE} 
\end{subfigure}
\hspace{-0.5em}
\begin{subfigure}[b]{0.25\textwidth}
 \includegraphics[width=\textwidth]{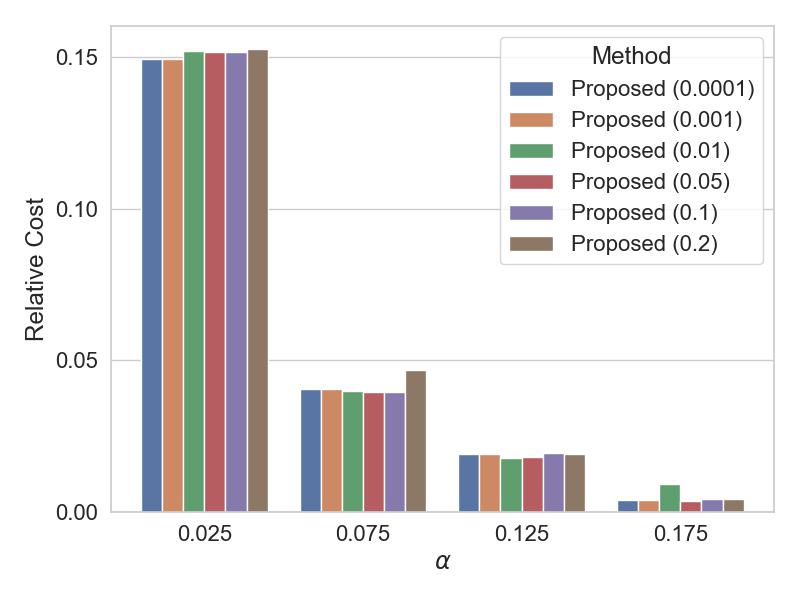}
\caption{VAE} 
\end{subfigure}
\end{center}
\caption{\rev{Influence of $\delta'$. Showing the scores of the free objective for different values of $\delta'$, which controls the width of the region of interest, defined in~\eqref{eq:region_i}}\looseness=-1}
\label{fig:res_delta_prime}
\end{figure}

\rev{\newpar{Influence of $\delta'$} We examine the influence of $\delta'$, which determines the boundaries of the region of interest. Figure.~\ref{fig:res_delta_prime} shows the scores obtained for different values of $\delta'$. We observe that in most cases there is no noticeable difference in the performance with respect to $\delta'$, indicating that our method is generally insensitive to this choice.}

\begin{figure}[t]
\begin{center}
\begin{subfigure}[b]{0.25\textwidth}
 \includegraphics[width=\textwidth]{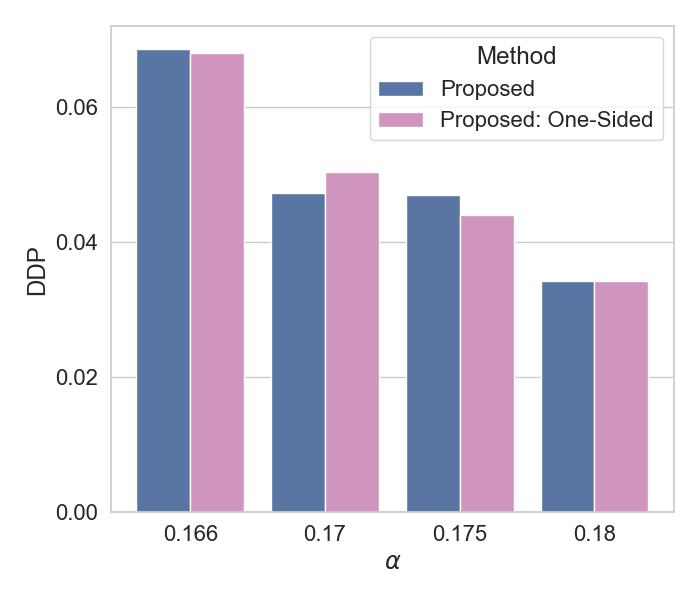}
\caption{Fairness}
\end{subfigure}
\hspace{-0.5em}
\begin{subfigure}[b]{0.25\textwidth}
 \includegraphics[width=\textwidth]{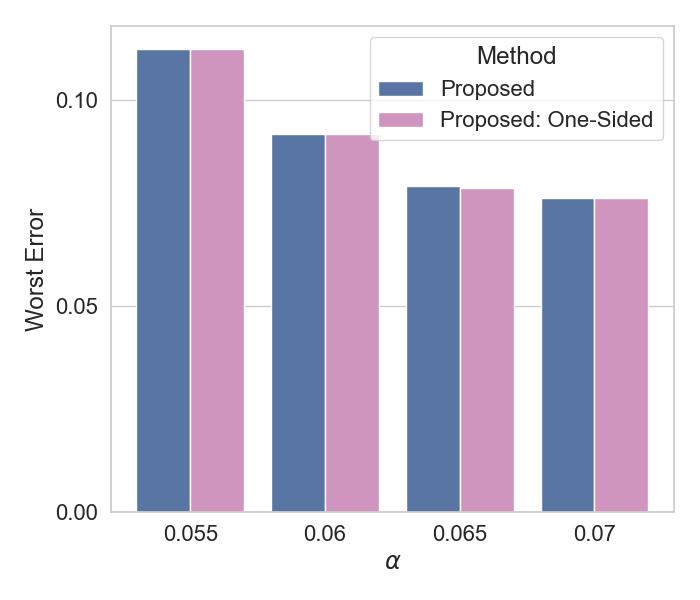}
\caption{Robustness}
\end{subfigure}
\hspace{-0.5em}
\begin{subfigure}[b]{0.25\textwidth}
 \includegraphics[width=\textwidth]{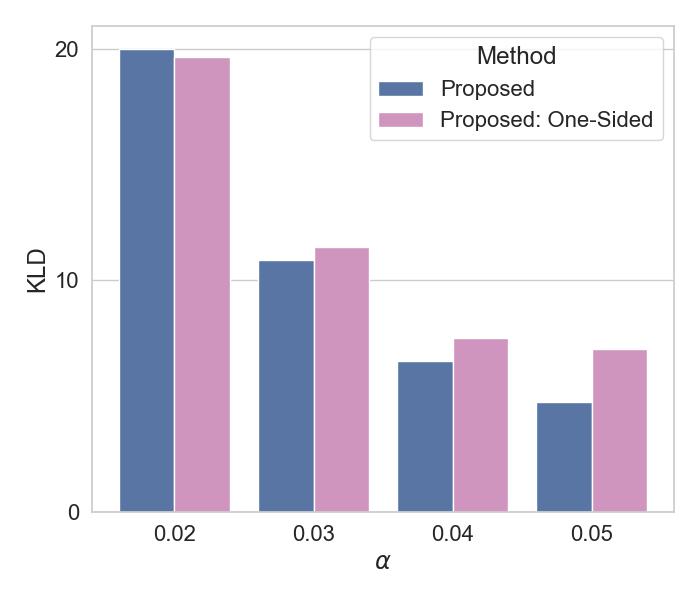}
\caption{VAE} 
\end{subfigure}
\hspace{-0.5em}
\begin{subfigure}[b]{0.25\textwidth}
 \includegraphics[width=\textwidth]{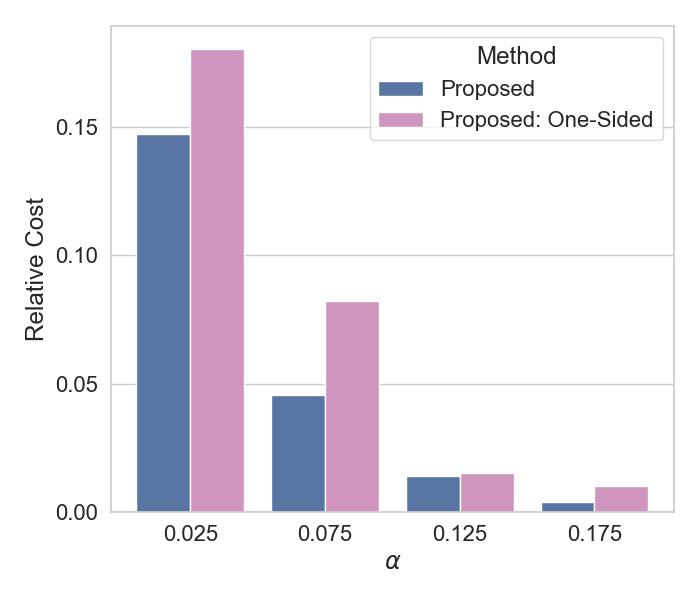}
\caption{Pruning} 
\end{subfigure}
\end{center}
\caption{\rev{Ablation study - one sided bound instead of the defined two-sided region. Presenting the scores obtained for the free objective.}\looseness=-1}
\label{fig:res_sided}
\end{figure}

\rev{\newpar{One sided bound} We compare the proposed method to the case that the \ac{BO} search is constrained by a one-sided bound at the upper limit defined in~\eqref{eq:region_i}. Fig.~\ref{fig:res_sided} shows the values of the free objective across tasks. We see that in most cases performing the search in the defined region of interest is preferable over a single-sided bound. This shows the benefit of removing low risk, inefficient configurations from the search space (the green section in Fig.~\ref{fig:explain}).} 

\newpage
\section{Algorithms}
Our proposed guided~\ac{BO} procedure and the overall method, are summarized in Algorithms~\ref{alg:bo}, and~\ref{alg:selection}, respectively.

\definecolor{darkgreen}{rgb}{0.31, 0.47, 0.26}
\begin{figure}[!t]
\centering
\begin{minipage}[t]{1\linewidth}
{\footnotesize
\begin{algorithm}[H]
\caption{Testing-Guided Bayesian Optimization}
\label{alg:bo}
{
\textbf{Definitions:} $\ell_1,\ldots,\ell_{c}$ and $\ell_\textrm{free}$ are the objective functions, $g_1,\ldots,g_{c}$ and $g_\textrm{free}$ are their associated surrogate models. $\ell_1^\textrm{low},\ldots,\ell_c^\textrm{low}$ and $\ell_1^\textrm{high},\ldots,\ell_c^\textrm{high}$ are the lower and upper bounds, respectively, for the first $c$ objectives. $\mathcal{C}_0=\{\bm{\lambda}_0,\ldots,\bm{\lambda}_{N_0}\}$ is an initial pool of configurations and $\mathcal{L}_0=\{\bm{\ell}(\bm{\lambda}_1),\ldots,\bm{\ell}(\bm{\lambda}_{N_0})\}$ are the associated objectives. $N$ is our total budget.\looseness=-1}
\begin{algorithmic}[1]
\Function{BO}{$\bm{\ell}$, $\mathcal{C}_0$, $\mathcal{L}_0$, $\{\ell_1^\textrm{low},\ldots,\ell_c^\textrm{low}\}$, $\{\ell_1^\textrm{high},\ldots,\ell_j^\textrm{high}\}$, $N$}
\State{$N_\mathrm{max}\leftarrow N-N_0$}
\State $\mathbf{r} \leftarrow \left(\ell_1^{\textrm{high}},\ldots,\ell_c^{\textrm{high}},\> \max_{\bm{\lambda}\in \mathcal{C}_0}\ell_{\textrm{free}}(\bm{\lambda})\right)$ \textcolor{darkgreen}{\Comment{\footnotesize Initialize reference point.}}
\For{$n = 0, 1, 2, \ldots, N_\mathrm{max}-1$}
    \State Fit $\bm{\hat{g}}$ on ($\mathcal{C}_n$, $\mathcal{L}_n$) 
    \textcolor{darkgreen}{\Comment{\footnotesize Fit surrogate models.}}
    \State $r_{c+1} \leftarrow \min_{\boldsymbol{\lambda}\in R^{\textrm{low}}}\hat{g}_\textrm{free}(\boldsymbol{\lambda})$ \textcolor{darkgreen}{\Comment{\footnotesize Update reference point.}}
    \State $\hat{\mathcal{P}}\leftarrow \textrm{ParetoFront}(\mathcal{L}_n)$ \textcolor{darkgreen}{\Comment{\footnotesize Filter Pareto front.}}
    \State{$\bm{\lambda}_{n+1}=\argmax_{\bm{\lambda}}HVI(\hat{\bm{g}}(\bm{\lambda}),\hat{\mathcal{P}};\mathbf{r})$.} \textcolor{darkgreen}{\Comment{\footnotesize Optimize acquisition function.}}
    \State{Evaluate $\bm{\ell}(\bm{\lambda}_{n+1})$} \textcolor{darkgreen}{\Comment{\footnotesize Evaluate new configuration.}}
    \State{$\mathcal{C}_{n+1} \leftarrow \mathcal{C}_n \cup \bm{\lambda}_{n+1}$.} \textcolor{darkgreen}{\Comment{\footnotesize Add new configuration.}}
    \State{$\mathcal{L}_{n+1} \leftarrow \mathcal{L}_n \cup \bm{\ell}(\bm{\lambda}_{n+1})$.} \textcolor{darkgreen}{\Comment{\footnotesize Add new objective values.}}
\EndFor
\State{$\mathcal{C}^{\textrm{BO}}\leftarrow \mathcal{C}_{N_\textrm{max}}$}
\State{\Return{$\mathcal{C}^{\textrm{BO}}$}}
\EndFunction
\end{algorithmic}
\end{algorithm}
}
\end{minipage}
\vspace{-20pt} 
\end{figure}

\begin{figure}[t]
\centering
\begin{minipage}{1\linewidth}
\begin{algorithm}[H]
\small
\caption{\small Configuration Selection}\label{alg:selection}
{\small \textbf{Definitions:} 
$f$ is a configurable model set by an hyperparameter $\bm{\lambda}$. $\mathcal{D}_\textrm{val}=\{X_i,Y_i\}_{i=1}^k$ and $\mathcal{D}_\textrm{cal}=\{X_i,Y_i\}_{i=k+1}^{k+m}$ are two disjoint subsets of validation and calibration data, respectively. $\{\ell_1, \ldots, \ell_{c}\}$ are constrained objective functions, and $\ell_\textrm{free}$ is a free objective. $ \{\alpha_1, \ldots, \alpha_{c}\}$ are user-specified bounds for the constrained objectives. $\Lambda$ is the configuration space. $\delta$ is the tolerance. $N$ is the optimization budget. \textsc{ParetoOptimalSet} returns Pareto optimal points.\looseness=-1
}
\begin{algorithmic}[1] 
\footnotesize
\vspace{2pt}
\Function{select}{$\mathcal{D}_\textrm{val}, \mathcal{D}_\textrm{cal}, \Lambda, \{\alpha_1, \ldots, \alpha_{c}\}, \delta, N$}  
\State Compute $\ell^\textrm{low}_i,\ell^\textrm{high}_i$ for $i\in \{1,\ldots,c\}$ based on~\eqref{eq:region_i} and~\eqref{eq:region} \textcolor{darkgreen}{\Comment{\footnotesize Determine the region of interest.}}
\State $\mathcal{C}_0, \mathcal{L}_0 \leftarrow$ Randomly sample an initial pool of configurations \textcolor{darkgreen}{\Comment{\footnotesize Generate an initial pool.}}
\State $\mathcal{C}^\textrm{BO} \leftarrow$ \textsc{BO}(\rev{$\mathcal{D}_\textrm{val}$},$\bm{\ell}$, $\mathcal{C}_0$, $\mathcal{L}_o$, $\{\ell_1^\textrm{low},\ldots,\ell_c^\textrm{low}\}$, $\{\ell_1^\textrm{high},\ldots,\ell_j^\textrm{high}\}$, $N$) \textcolor{darkgreen}{\Comment{\footnotesize \ac{BO} via Algorithm~\ref{alg:bo}.}}
\State $\mathcal{C}^\textrm{p}\leftarrow $\textsc{ParetoOptimalSet}($\mathcal{C}^\textrm{BO}$) \textcolor{darkgreen}{\Comment{\footnotesize Filter Pareto points.}}
\State Compute $p^\textrm{val}_{\bm{\lambda}}$ over $\mathcal{D}_\textrm{val}$ for all $\bm{\lambda} \in \mathcal{C}^\textrm{p}$ \textcolor{darkgreen}{\Comment{\footnotesize Compute approximated p-values.}}
\State $\mathcal{C}^\textrm{o}\leftarrow$ Order configurations according to increasing $p^\textrm{val}_{\bm{\lambda}}$ \textcolor{darkgreen}{\Comment{\footnotesize Order configurations.}}
\State Compute $p^\textrm{cal}_{\bm{\lambda}}$ over $\mathcal{D}_\textrm{cal}$ for all $\bm{\lambda} \in \mathcal{C}^\textrm{o}$ \textcolor{darkgreen}{\Comment{\footnotesize Compute p-values.}}
\State Apply \ac{FST}: $\mathcal{C}^\textrm{valid} = \{\bm{\lambda}^{(j)}: j< J \}, \>\> J = \min_j \{j: p^\textrm{cal}_{\bm{\lambda}}\geq \delta\}$ \textcolor{darkgreen}{\Comment{\footnotesize Apply \ac{FST}.}}
\State $\bm{\lambda}^*=\min_{\bm{\lambda}\in \mathcal{C}^\textrm{valid}}\ell_{\textrm{free}}(\bm{\lambda})$ \textcolor{darkgreen}{\Comment{\footnotesize Choose best-performing configuration.}} \\
\hspace*{\algorithmicindent}\Return  $\bm{\lambda}^*$ 
\EndFunction
\end{algorithmic}
\end{algorithm}
\end{minipage}
\vspace{-12pt}
\end{figure}
\section{Mathematical Details}
\label{sec:math_details}
\subsection{\rev{Derivation of the Region of Interest}}
\rev{Suppose the loss is bounded above by 1, then Hoeffding's inequality~\cite{hoeffding1994probability} is given by:
\begin{equation}
P\left( \hat{\ell}(\bm{\lambda}) - \ell(\bm{\lambda}) \le -t\right) \le e^{-2nt^2}.
\label{eq:hoeffding_bound_a}
\end{equation}
and 
\begin{equation}
P\left( \hat{\ell}(\bm{\lambda}) - \ell(\bm{\lambda}) \ge t\right) \le e^{-2nt^2}.
\label{eq:hoeffding_bound_b}
\end{equation}
for $t>0$. Taking $u=e^{-2nt^2}$, we have $t= \sqrt{\frac{\log\left(1/u\right)}{2n}}$, hence:
\begin{equation}
P\left( \hat{\ell}(\bm{\lambda}) - \ell(\bm{\lambda}) \le - \sqrt{\frac{\log\left(1/u\right)}{2n}}\right) \le u.
\label{eq:tranformed_hoeffding_a}
\end{equation}
and
\begin{equation}
P\left( \hat{\ell}(\bm{\lambda}) - \ell(\bm{\lambda}) \ge  \sqrt{\frac{\log\left(1/u\right)}{2n}}\right) \le u.
\label{eq:tranformed_hoeffding_b}
\end{equation}
This implies an upper confidence bound 
\begin{equation}
{\ell}^+_\textrm{HF}(\bm{\lambda}) = \hat{\ell}(\bm{\lambda}) + \sqrt{\frac{\log\left(1/u\right)}{2n}}
\label{eq:upper_bound}
\end{equation}
and a lower confidence bound
\begin{equation}
{\ell}^{-}_\textrm{HF}(\bm{\lambda}) = \hat{\ell}(\bm{\lambda}) - \sqrt{\frac{\log\left(1/u\right)}{2n}}.
\label{eq:lower_bound}
\end{equation}}

\rev{In addition, we can use Hoeffding's inequality to derive a valid p-value under the null hypothesis $H_{\bm{\lambda}}:  \ell(\bm{\lambda}) > \alpha$.
By~\eqref{eq:tranformed_hoeffding_a}, we get:
\begin{equation}
P\left( \hat{\ell}(\bm{\lambda}) - \alpha \le -\sqrt{\frac{\log\left(1/u\right)}{2n}}\right)\le P\left( \hat{\ell}(\bm{\lambda}) - \ell(\bm{\lambda}) \le -\sqrt{\frac{\log\left(1/u\right)}{2n}}\right) \le u.
\label{eq:insert_alpha}
\end{equation}
For $\hat{\ell}(\bm{\lambda})<\alpha$, we rearrange~\eqref{eq:insert_alpha} to obtain:
\begin{equation}
P\left(e^{-2n(\alpha-\hat{\ell}(\bm{\lambda}))^2}\leq u\right) \le u,
\label{eq:tranformed_hoeffding_a2}
\end{equation}
which implies that $p_{\bm{\lambda}}^\textrm{HF}\coloneqq e^{-2 m\left(\alpha-\hat{\ell}(\bm{\lambda})\right)^2_+}$ is super-uniform, hence is a valid p-value. Comparing $p_{\bm{\lambda}}^\textrm{HF}$ to $\delta$, yields the maximum empirical loss $\hat{\ell}(\bm{\lambda})$, evaluated over a calibration set of size $m$, which can pass the test with significance level $\delta$:
\begin{equation}
\alpha^\textrm{max}=\alpha-\sqrt{\frac{\log\left(1/\delta\right)}{2m}}.
\end{equation}
This can be equivalently obtained from the upper bound~\eqref{eq:upper_bound}.}  

A tighter alternative to Hoeffding p-value was proposed in~\cite{bates2021distribution} based Hoeffding and Bentkus inequalities. The Hoeffding-Bentkus p-value is given by:
\begin{equation}
p^{\textrm{HB}}_{\bm{\lambda}} = \min\left(\exp\{-mh_1(\hat{\ell}(\bm{\lambda})\wedge\alpha,\alpha)\},e\mathbb{P}\left(\textrm{Binom}(m,\alpha)\leq\lceil m \hat{\ell}(\bm{\lambda})\rceil\right)\right)  
\label{eq:HB_pval}
\end{equation}
where $h_1(a,b)=a\log(\frac{a}{b})+(1-a)\log(\frac{1-a}{1-b})$.
Note that for a given $\delta$ we can numerically extract from~\eqref{eq:HB_pval} the upper and lower bounds corresponding to a $1-\delta$ confidence interval, and use it to define the region of interest as in~\eqref{eq:region_i}.

\subsection{Proof of Proposition~\ref{thm:ltt}}
\rev{The proof is based on~\cite{angelopoulos2021learn,laufer2023efficiently}, which we repeat here for completeness.}

\begin{proof}
\rev{Recall that $\mathcal{D}_{\mathrm{val}}$ and $\mathcal{D}_{\mathrm{cal}}$ are two disjoint, i.i.d. datasets. Therefore, $\mathcal{D}_{\mathrm{cal}}$ is i.i.d. w.r.t the returned configuration set optimized in Algorithm~\ref{alg:bo} over $\mathcal{D}_{\mathrm{val}}$.}

\rev{We now prove that the testing procedure returns a set of valid configurations with \ac{FWER} bounded by $\delta$.
Let $H_{\bm{\lambda}'}$ be the first true null hypothesis in the sequence. Given that $p_{\bm{\lambda}'}$ is a super uniform p-value under $H_{\bm{\lambda}'}$, the probability of making a false discovery at $\bm{\lambda}'$ is bounded by $\delta$. This means that the event that  $H_{\bm{\lambda}'}$ is rejected (false discovery) occurs with probability lower than $\delta$. According to the sequential testing procedure, all other $H_{\bm{\lambda}}$ that follow are also rejected (regardless of if $H_{\bm{\lambda}}$ is true or not). Therefore the probability of making any false discovery is bounded by $\delta$, which satisfies the \ac{FWER} control requirement.} 

\end{proof}

\subsection{Hypervolume}
\label{sec:hypervolume}
\begin{figure}[H]
\begin{center}
\includegraphics[width=0.5\textwidth]{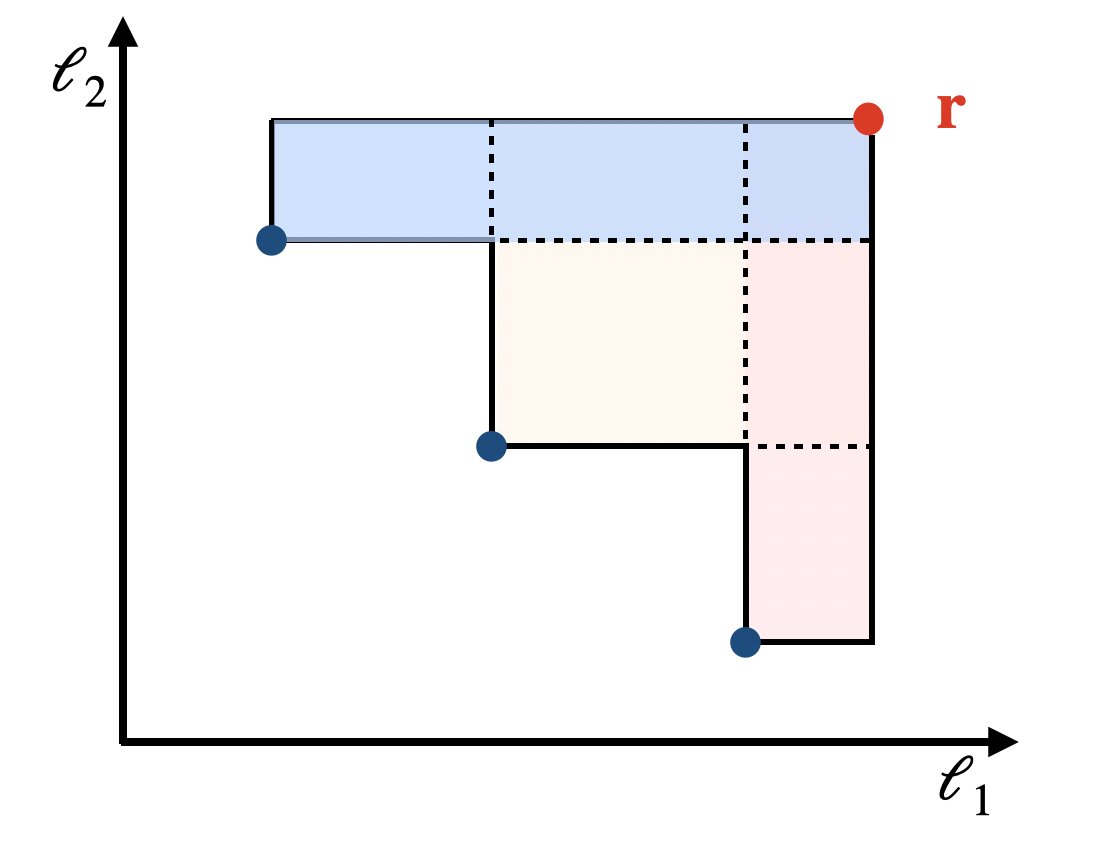}
\end{center}
\caption{
An illustration of the hypervolume in the 2-dimensional case. The reference point is marked in red and three Pareto optimal points are marked in blue.\looseness=-1}
\label{fig:hypervolume}
\end{figure}

An illustration of the hypervolume defined 
in~\eqref{eq:hypervolume} is given in Fig.~\ref{fig:hypervolume} for the 2-dimensional case.
It can be seen that the hypervolume is equivalent to the volume of the union of the boxes created by the Pareto optimal points.

\subsection{\rev{Illustration}}
\begin{figure}[!t]
\begin{center}
\includegraphics[width=0.99\textwidth]{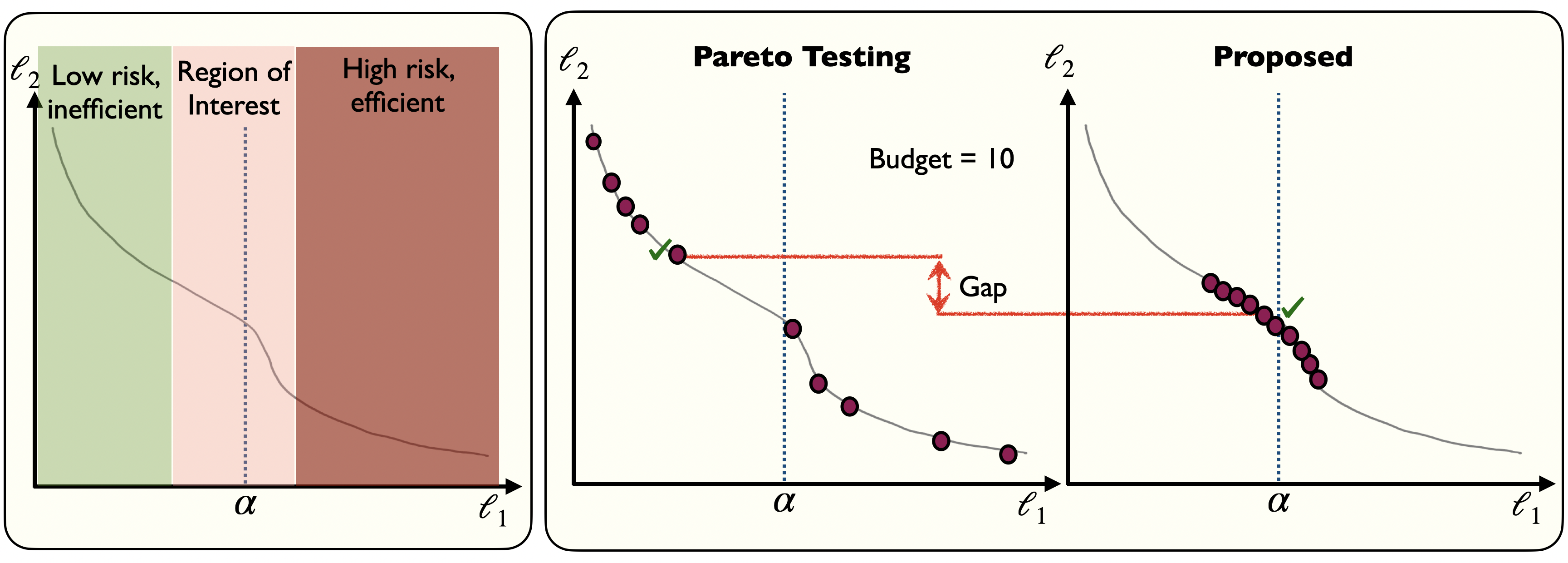}
\end{center}
\vspace{-0.4cm} 
\caption{
\rev{Left: Illustration of the different parts of the Pareto front. The green region consists of configurations that are low risk ($\ell_1\ll\alpha$) but inefficient in terms of the free objective $\ell_2$. The brown region consists of configurations that are efficient but high risk ($\ell_1\gg\alpha$) and cannot pass the test. In the middle we define the region of interest containing configurations that are likely to be both valid and efficient. Right: comparing the proposed method to Pareto Testing for optimization budget $N=10$. In Pareto Testing there is no control on the distribution of the configurations on the front, while our method focuses on the region of interest. As a result, there is a gap in the minimization of $\ell_2$ for the chosen valid configuration (marked by v) in favor of the proposed method.} \looseness=-1} 
\label{fig:explain}
\vspace{-5pt}
\end{figure}

\rev{Figure~\ref{fig:explain} shows the partition of the Pareto front to the different regions, and demonstrates the difference between Pareto Testing and the proposed method.}

\end{document}